\documentclass[letterpaper,journal]{IEEEtran}
\usepackage{amsmath,amsfonts}
\usepackage{array}
\usepackage{textcomp}
\usepackage{stfloats}
\usepackage{url}
\usepackage{verbatim}
\usepackage{graphicx}
\usepackage{cite}
\usepackage[table]{xcolor}
\usepackage{colortbl}
\colorlet{bestgray}{gray!25}
\usepackage{mathtools}
\usepackage{amssymb}
\usepackage{tabulary}
\usepackage{booktabs}
\usepackage[hidelinks,hyperfootnotes=false]{hyperref}
\usepackage[bottom]{footmisc}
\hypersetup{
  pdftitle={Robust Active-Perception Control for Global-State-Free Aerial-Ground Cooperation},
  pdfauthor={Mingxuan Zhang, Jiajun Yu, Baozhe Zhang, Pengxiang Zhou, Wentao Liu, Fei Gao, Chao Xu, and Yanjun Cao},
  pdfkeywords={Nonlinear MPC, motion control, vision-based navigation, global-state-free control}
}
\usepackage{setspace}
\usepackage{bm}
\usepackage{multirow}
\usepackage{makecell}
\usepackage{adjustbox}
\usepackage{pifont}

\newcommand{\B}[1]{\boldsymbol{\mathbf{#1}}}

\newcommand{\skewsym}[1]{[#1]_{\times}}

\newcommand{\nbx}[1]{{}^N{#1}_{B}}
\newcommand{\nnx}[1]{{}^N{#1}_{N}}
\newcommand{\bbx}[1]{{}^B{#1}_{B}}

\newcommand{\rotmat}[2]{{}^{#1}_{#2}\B{R}}

\title{
Robust Active-Perception Control for Global-State-Free Aerial-Ground Cooperation
\thanks{\textsuperscript{\textdagger}Indicates equal contribution. This work was partially supported by the National Natural Science Foundation of China (Grant No. 62636012) and the Key R\&D Project of China National Tobacco Corporation (Grant No. 110202402018). (Corresponding authors: Yanjun Cao, Chao Xu.)}
}

\author{Mingxuan Zhang\textsuperscript{\textdagger}, Jiajun Yu\textsuperscript{\textdagger}, Baozhe Zhang, Pengxiang Zhou, Wentao Liu, Fei Gao, Chao Xu, and Yanjun Cao%
\thanks{Mingxuan Zhang, Jiajun Yu, Pengxiang Zhou, Wentao Liu, Fei Gao, Chao Xu, and Yanjun Cao are with the State Key Laboratory of Industrial Control
Technology, Institute of Cyber-Systems and Control, Zhejiang University,
Hangzhou 310027, China, and also with the Huzhou Institute of Zhejiang University and
Huzhou Key Laboratory of Autonomous Systems, Huzhou 313000, China (email: mixianz@zju.edu.cn; yanjunhi@zju.edu.cn).

Baozhe Zhang is with The Chinese University of Hong Kong, Shenzhen 518172, China (email: baozhezhang@link.cuhk.edu.cn).}%
}

\begin{document}

\maketitle

\begin{abstract}

Aerial-ground cooperation requires real-time UAV--UGV relative-state information. Instead of maintaining global estimates for both robots, direct control in a UGV-attached non-inertial frame avoids reliance on global localization. Vision-based relative pose estimation with a passive marker offers a low-cost and effective solution. 
However, a fixed camera may lose sight of the moving UGV when the required UAV attitude conflicts with the field-of-view (FOV) constraint. To address this, we propose COPA, a robust active-perception framework for global-state-free aerial-ground cooperation. We use a single-axis gimbal to decouple the camera optical axis from the UAV pitch attitude. We derive an active-perception model that relates UAV motion, gimbal angle, and UGV motion to the target image-plane state. 
A Temporal Convolutional Network (TCN) predicts short-horizon UGV acceleration and angular velocity from recent motion history without global-state measurements. The model predictive control (MPC) uses these predictions to jointly optimize UAV and gimbal control. Simulations show that COPA maintains continuous target visibility, while ablation studies confirm that the TCN reduces peak errors during UGV motion transitions.
Real-world experiments with UGV accelerations up to $3\,\mathrm{m/s^2}$ and yaw rates up to $1.0\,\mathrm{rad/s}$ demonstrate robust tracking.

\end{abstract}


\section{Introduction}

\IEEEPARstart{A}{erial-ground} collaboration has been studied for cooperative target tracking, heterogeneous motion coordination, and autonomous landing on moving ground vehicles \cite{yu2015cooperative,mu2018distributed,lim2022hemispherical}. In these tasks, the robots must coordinate their actions, which requires knowing each other's relative state. A common approach is for each robot to estimate its own pose in a world frame using onboard localization \cite{chen2023stereo,yang2024mcov}. The relative state is then recovered by combining the two world-frame poses. However, this method couples the cooperative task to the state estimations of all robots, with each bearing the cost of maintaining a global estimation. This motivates direct UAV control in the target-attached frame. Since the target accelerates, decelerates and rotates, the frame is non-inertial.

\begin{figure}[!t]
\centering
\includegraphics[width=\columnwidth]{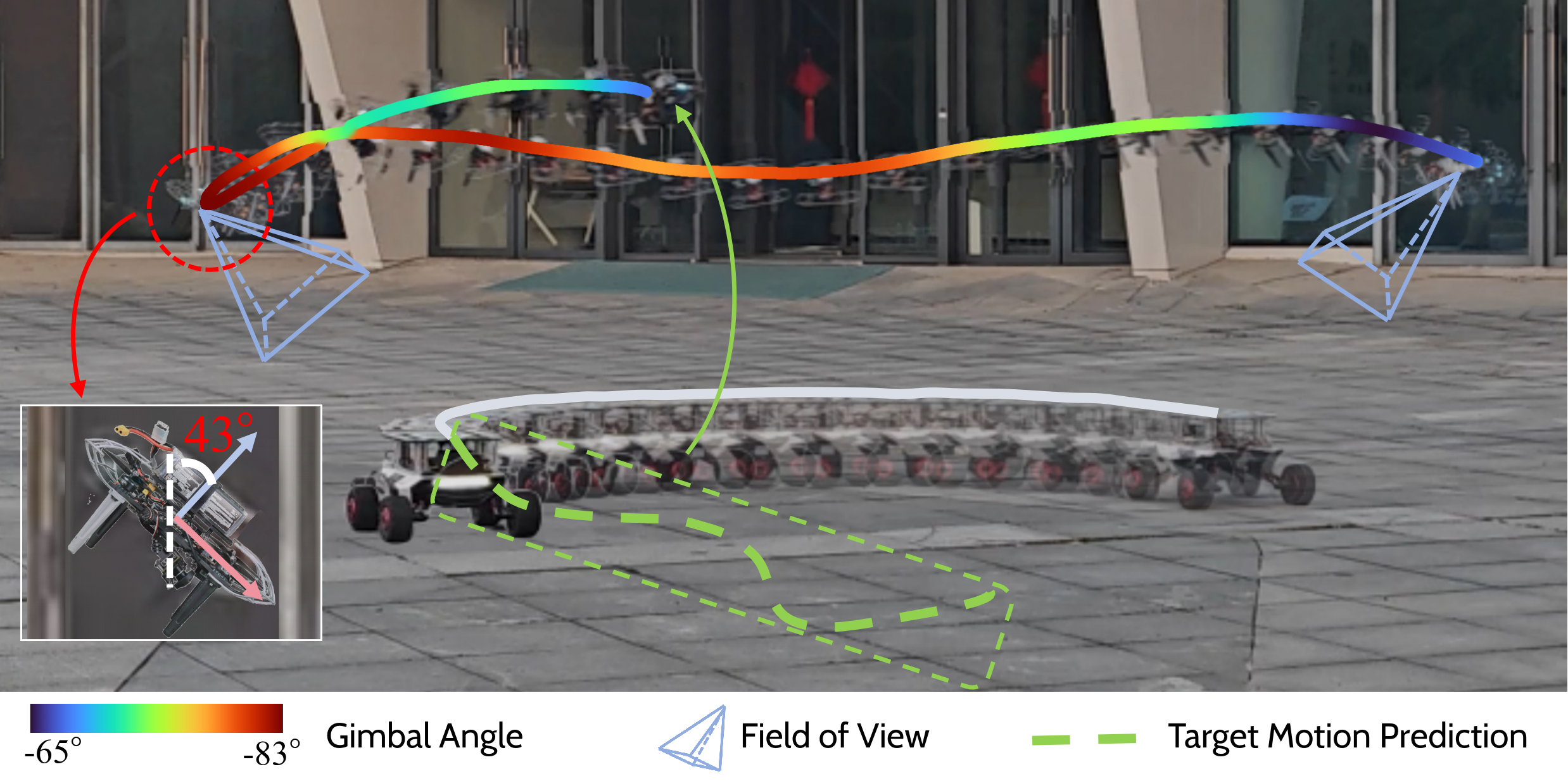}
\caption{Real-world illustration of COPA under dynamic UGV motion. UAV motion is colored by gimbal angle, representative camera frustums show the FOV, and the dashed green curve denotes target-motion prediction. The inset shows the UAV body pitch reaching $43^\circ$ and the corresponding camera FOV.}
\label{fig:teaser}
\end{figure}

Relative motion control in a non-inertial frame has a long history in aerospace \cite{sun20156} and has been applied to aerial-ground cooperation recently \cite{zhang2023coni}. Implementing such control requires reliable real-time estimations of the relative state. A single ultra-wideband (UWB) range measurement provides distance only \cite{guo2020ultrawideband} and recovering full relative pose generally requires multiple anchors, antenna arrays, or additional sensing\cite{CREPES}. AprilTag-based vision methods \cite{apriltag,krogius2019flexible} require only a passive marker on the UGV and give the full 6-DOF relative pose from a single visual detection.
However, maintaining the FOV constraint in the non-inertial frame is challenging. A body-fixed camera introduces a conflict between target visibility and the UAV attitude required for tracking. Related methods \cite{falanga2018pampc,jacquet2020perception,penin2017vision,penin2018vision} incorporate perception quality into the control but cannot eliminate this coupling.

In the non-inertial frame, the relative dynamics depend on target-motion parameters, including acceleration and angular velocity. However, the future evolution of these time-varying parameters is unknown when generating UAV commands, which presents another challenge. A simplification \cite{zhang2023coni,zhang2025global} is to keep the latest measurements unchanged over a short future interval. This assumption is reasonable when the UGV moves steadily, but becomes inaccurate during acceleration, deceleration, or sharp turns.
The resulting mismatch prevents the UAV from responding promptly to changes in target motion, producing large transient errors.
Predicting short-term UGV motion allows the UAV to adjust its motion in advance.

To address these challenges, we propose COPA (Cooperative Non-Inertial Perception-Aware Framework), a robust active-perception framework using a single-axis gimbal for global-state-free aerial-ground cooperation. 
Specifically, we mount a single-axis gimbal on the UAV to decouple the camera optical axis from UAV pitch motion.
Based on this design, we derive an active-perception model in the non-inertial frame. The model describes how UAV motion, gimbal angle, and UGV motion affect the target's image-plane position and velocity. 
Since the non-inertial dynamics depend on target motion parameters, including the UGV acceleration and angular velocity, a Temporal Convolutional Network (TCN) predicts their short-horizon evolution from a short window of UGV motion history.
The predicted parameters are then incorporated into a unified MPC formulation, which jointly optimizes UAV flight control and gimbal angle control.

Fig.~\ref{fig:teaser} shows the performance of COPA in a real-world experiment. This joint formulation maintains target visibility and relative trajectory tracking. The look-ahead prediction enables anticipatory UAV and gimbal adjustments, reducing transient tracking errors and image-plane deviation.

The contributions of this work are:
\begin{itemize}
    \item We propose an active-perception MPC in the non-inertial frame. 
    It jointly optimizes the UAV flight control and gimbal angle control to maintain visibility and relative tracking.

    \item We design a learning-based prediction module that predicts short-horizon target motion parameters from UGV motion history without global state measurement.  
    It reduces peak tracking error during motion transitions and improves the overall tracking performance.

    \item We demonstrate the effectiveness and robustness of proposed vision-based UAV--UGV collaborative system in the simulation and real-world experiments.
\end{itemize}

\section{Related Work}

\subsection{Vision-Based Aerial-Ground Cooperation}

Aerial target tracking from a UAV forms the perceptual foundation of aerial-ground cooperation. Fast-Tracker \cite{han2021fast} combined polynomial target motion prediction with kinodynamic trajectory optimization for tracking agile ground targets in cluttered environments. Elastic Tracker \cite{ji2022elastic} further embedded occlusion-aware path finding to maintain continuous visibility. Both systems rely on world-frame localization and a body-fixed camera, requiring the UAV to redirect its entire body to keep the target in view.

Vision-based systems are constrained by the camera's field of view. Adjusting the yaw angle provides a direct way to reorient the camera toward the target. RAPTOR \cite{zhou2021raptor} first refines the flight trajectory to expose potentially dangerous unknown regions and then plans yaw to maximize information gain and smoothness. For aerial videography, Auto Filmer \cite{zhang2022auto} jointly optimizes the quadrotor trajectory and the yaw angle of a single-axis gimbal to maintain the desired image composition while partially decoupling camera heading from vehicle attitude. However, yaw-only camera actuation does not remove the coupling in the vertical image direction during aggressive pitch maneuvers.

These works approach the perception-control problem in a relative frame. NOVA \cite{saviolo2025nova} and HUNT \cite{saviolo2025hunt} formulate perception, estimation, and control entirely in the target's reference frame using only onboard stereo vision and IMU. Paris et al. \cite{paris2020dynamic} combine AprilTag-based relative pose estimation with receding-horizon MPC for precise landing. The cited systems predominantly use body-fixed cameras or camera actuation that does not explicitly model the pitch-induced FOV coupling considered here.

COPA addresses this with a single-axis servo gimbal at the hardware level and a unified active-perception MPC in the non-inertial frame at the software level.

\subsection{Target-Motion Prediction for Predictive Control}

MPC performance depends on the accuracy of the prediction model.
A common strategy is to learn this model from data. 
Most such work targets the ego-vehicle's own dynamics.
Torrente et al.~\cite{torrente2021data} learn residual aerodynamic effects with Gaussian processes and embed them in an MPC, reducing tracking error at high speeds.
Salzmann et al.~\cite{salzmann2023realtime} integrate large neural-network dynamics models into a real-time MPC for agile flight.
KNODE-MPC~\cite{chee2022knode} augments a first-principles model with a neural ordinary differential equation, and Saviolo et al.~\cite{saviolo2022physics} learn quadrotor dynamics with a physics-inspired temporal convolutional network.
Kumar et al.~\cite{kumar2021rma} adapt the control policy online from a history of ego states.
In every case, the learned component describes the ego-vehicle itself. 

COPA faces a different prediction problem.
In the non-inertial frame, the relative dynamics depend on the target-motion parameters, the UGV's acceleration and angular velocity.
These parameters are exogenous. 
They are set by the UGV's future motion, not by any UAV command.
Prior non-inertial control methods~\cite{zhang2023coni} hold these
parameters constant over the horizon.
This is accurate for steady motion but breaks down at transients such as acceleration, deceleration, or sharp turns.
COPA instead learns to forecast the parameter sequence.
Temporal convolutional networks predict multi-step robot motion in a single forward pass~\cite{looper2022temporal}. 
This matches the horizon-length prediction the MPC needs.
The same architecture appears in PI-TCN~\cite{saviolo2022physics}, but its role differs. 
PI-TCN learns the ego-vehicle's state-to-state dynamics. 
COPA forecasts an exogenous signal and feeds it to the MPC as time-varying parameters.
The resulting TCN-based Target-Motion Predictor~\cite{bai2018empirical} maps the UGV's recent motion history to horizon-length predictions of its acceleration and angular velocity, compensating for the model mismatch left by the constant-parameter assumption.

\section{Methodology}

\begin{figure}[!t]
\centering
\includegraphics[width=\columnwidth]{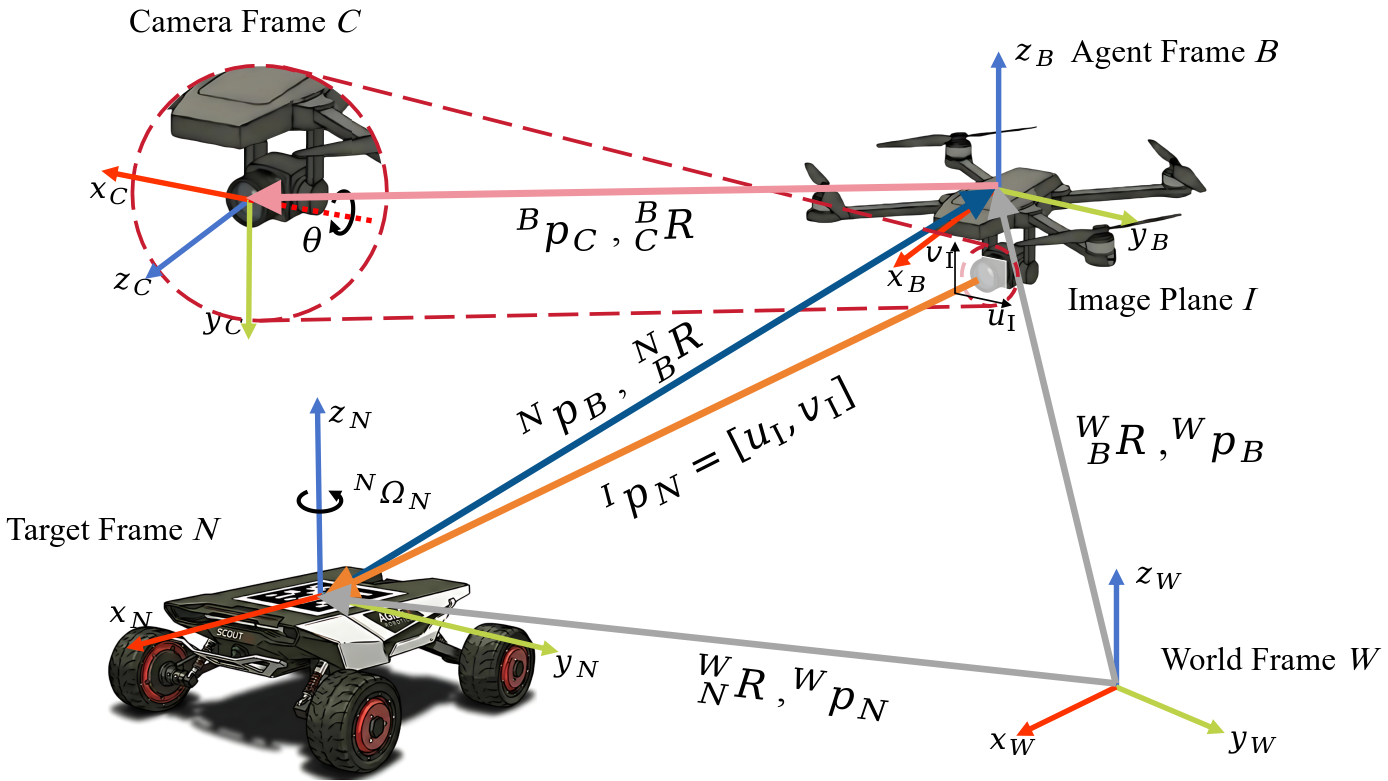}
\caption{The transformation relationships between the world frame ($W$), the agent frame ($B$), the target frame ($N$), the camera frame ($C$), and the image plane ($I$).}
\label{relative}
\end{figure}

\begin{figure*}[!t]
    \centering
    \includegraphics[width=0.95\textwidth]{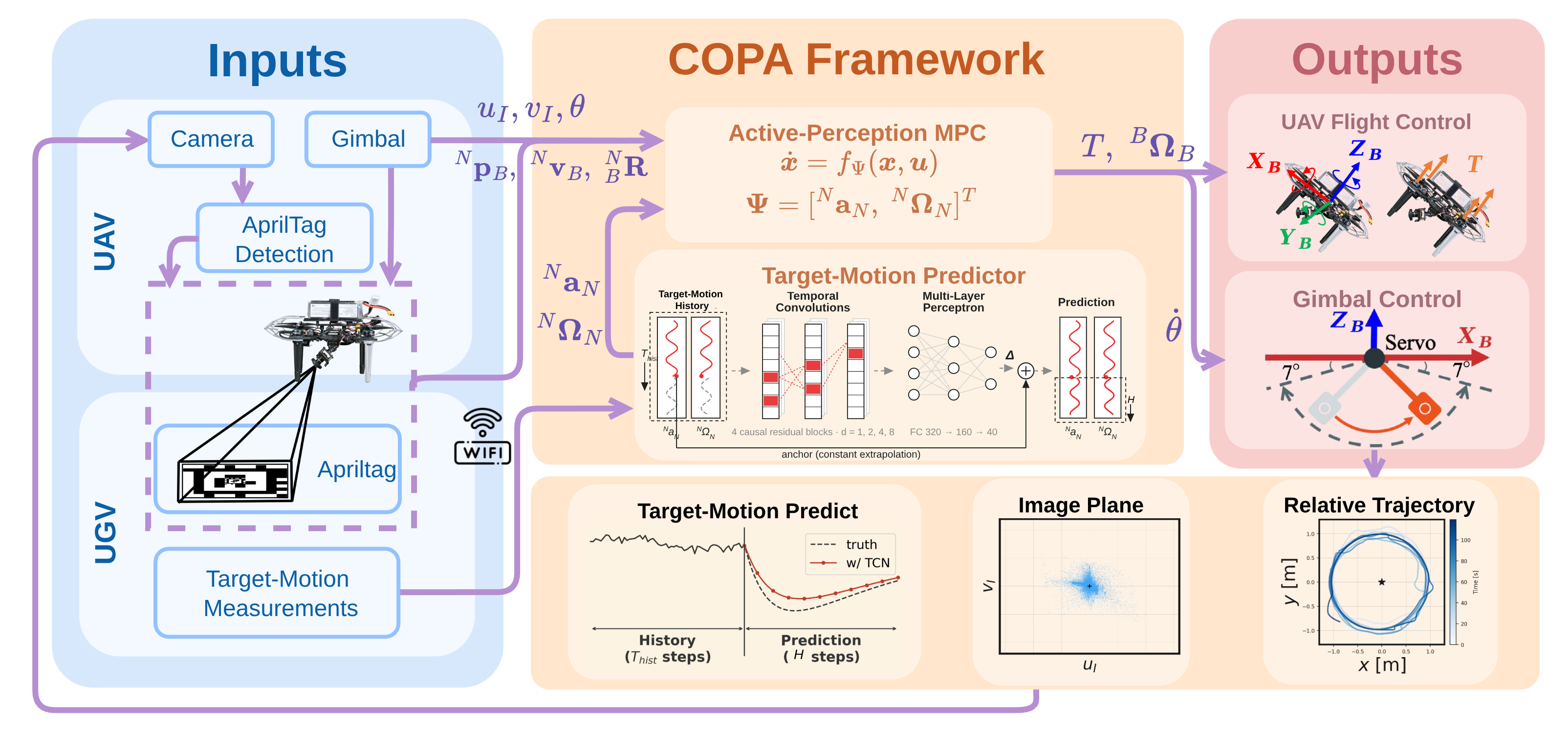}
    \caption{Overview of the proposed COPA system. AprilTag detection and gimbal sensing provide the relative state and image-plane measurements, while UGV motion estimates are transmitted to the UAV. The TCN-based Target-Motion Predictor forecasts the UGV acceleration and angular velocity over the MPC horizon. The Active-Perception MPC uses these predictions to jointly generate UAV flight and gimbal commands, maintaining target visibility and relative trajectory tracking. }
    \label{fig:overall}
\end{figure*}

The relationships among the frames used in our system are shown in Fig.~\ref{relative}. Frame $B$ is attached to the quadrotor body, frame $N$ is attached to the UGV body, and frame $W$ denotes the inertial world frame. The camera frame $C$ is attached to $B$ through a single-axis gimbal rotating by $\theta$. The target is observed in $C$ and projected onto the image plane $I$. Table~\ref{tab:notations} lists the main notations used in this paper.
Fig.~\ref{fig:overall} summarizes COPA's information flow from onboard data to UAV and gimbal commands.

\begin{table}[!t]
    \caption{Table of Notations}
    \label{tab:notations}
    \centering
    \renewcommand{\arraystretch}{1.25}
    \begin{tabular}{r c p{6.0cm}}
        \toprule
        ${}^{*}\B{p}_{\#}$               & $\triangleq$ & Position of frame $\#$ origin in frame $*$ \\
        ${}^{*}\B{v}_{\#}$               & $\triangleq$ & Velocity of frame $\#$ origin in frame $*$ \\
        ${}^{*}_{\#}\B{R}$               & $\triangleq$ & Rotation matrix from frame $\#$ to frame $*$ \\
        ${}^{*}\B{\omega}_{\#}^{\wedge}$ & $\triangleq$ & Angular velocity of frame $\#$ relative to frame $\wedge$, expressed in frame $*$ \\
        ${}^{*}\B{\Omega}_{*}$          & $\triangleq$ & Absolute angular velocity of frame $*$, expressed in frame $*$ \\     
        $\B{\Psi}$                      & $\triangleq$ & Target-motion parameter vector \\
        $\bbx{\B{T}}$                    & $\triangleq$ & Mass-normalized collective thrust of UAV \\
        $u_I,v_I$                      & $\triangleq$ & Target coordinates on the image plane \\
        $\theta,\;\dot{\theta}$          & $\triangleq$ & Gimbal pitch angle and angular velocity \\
        \bottomrule
    \end{tabular}
\end{table}

\subsection{Active-Perception Dynamics Model in the Non-Inertial Frame}
\label{sec:dynamics}

Motivated by prior work on relative dynamics in non-inertial frames \cite{zhang2023coni}, we derive an active-perception model that captures the coupled effects of UAV motion, gimbal actuation, and target motion.
The state vector of the model is defined as $\B{x} = [\nbx{\B{p}};\nbx{\B{v}};\rotmat{N}{B};u_I;v_I;\theta] \in \mathbb{R}^{6} \times \mathrm{SO}(3) \times \mathbb{R}^{3}$, and the control vector is $\B{u} = [T;\bbx{\B{\Omega}};\dot{\theta}] \in \mathbb{R}^{5}$. $T$ is the magnitude of the mass-normalized thrust vector $\bbx{\B{T}}=[0,0,T]^T$, and $\bbx{\B{\Omega}}$ is the UAV body rate. The relative translational dynamics are

\begin{equation}
    \begin{aligned}
        \label{eq:relative_dynamics}
        \nbx{\dot{\B{p}}} & = \nbx{\B{v}}                                                          \\
        \nbx{\dot{\B{v}}} & = -\skewsym{\nnx{\B{\beta}}} \nbx{\B{p}} - 2\skewsym{\nnx{\B{\Omega}}} \nbx{\B{v}}                                    \\
                          & -\skewsym{\nnx{\B{\Omega}}}^2 \nbx{\B{p}} + {}^N\B{g} + \rotmat{N}{B}\,\bbx{\B{T}} - \nnx{\B{a}},
    \end{aligned}
\end{equation}

where ${}^{N}\B{g}$ denotes gravity expressed in frame $N$. The controller adopts a reduced-order approximation that omits the target-frame angular-acceleration term $\nnx{\B{\beta}}$. This does not assume a constant yaw rate; \(\nnx{\B{\Omega}}\) is updated at every control cycle, and the target-motion predictor provides step-dependent angular-velocity predictions over the MPC horizon. This approximation is evaluated experimentally within the tested operating envelope. Accordingly, the target-motion parameter vector is defined as

\begin{equation}
    \B{\Psi} = [\nnx{\B{a}}; \nnx{\B{\Omega}}].
\end{equation}
These quantities can be estimated directly from onboard motion measurements without relying on any global-state information.

The time derivative of a rotation matrix is given by:
\begin{equation}
    \label{eq:rotation_derivative}
    \dot{\rotmat{X}{Y}}=[{}^X\B{\omega}_Y^X]_{\times} \:\rotmat{X}{Y}.
\end{equation}

The rotation matrix from B to N is:
\begin{equation}
    \rotmat{N}{B}=\rotmat{N}{W}\:\rotmat{W}{B}.
\end{equation}
Substituting Eq.~\eqref{eq:rotation_derivative}, we obtain the derivative of $\rotmat{N}{B}$
\begin{equation}
\begin{aligned}
    \label{R_N_B_DOT}
\dot{\rotmat{N}{B}}
&= \left[-\rotmat{N}{W}\,{}^W\B{\omega}_N^W\right]_{\times}\,\rotmat{N}{B} \,+\, \rotmat{N}{W}\,\left[\rotmat{W}{B}\,{}^B\B{\Omega}_B\right]_{\times}\,\rotmat{B}{W}\\[0.8em]
&= -\left[{}^N\B{\Omega}_N\right]_{\times}\, \rotmat{N}{B} \,+\, \rotmat{N}{B}\,\left[{}^B\B{\Omega}_B\right]_{\times}.
\end{aligned}
\end{equation}
This gives the kinematics of the attitude component $\rotmat{N}{B}$ in the system state.

Similarly, we obtain the derivative of \(\rotmat{C}{B}\)
\begin{equation}
\begin{aligned}
    \label{R_C_B_DOT}
    \dot{\rotmat{C}{B}} &= [{}^C\B{\omega}_B^C]_{\times}\,\rotmat{C}{B} = -\rotmat{C}{B}\,[{}^B\B{\omega}_C^B]_{\times}\, \rotmat{B}{C}\,\rotmat{C}{B} \\[0.8em]
    &= -\rotmat{C}{B}\,[\rotmat{B}{C}\,{}^C\B{\omega}_C^B]_{\times} = -[{}^C\B{\omega}_C^B]_{\times}\,\rotmat{C}{B}.
\end{aligned}
\end{equation}

$\nnx{\B{\Omega}}$ and $\bbx{\B{\Omega}}$ are the angular velocities of the UGV frame $N$ and UAV body frame $B$. ${}^C\B{\omega}_C^B$ is the angular velocity of frame $C$ relative to frame $B$, expressed in frame $C$. Because the gimbal has a single rotational degree of freedom, it is determined entirely by the gimbal rate:
\begin{equation}
    {}^C\B{\omega}_C^B
    =\begin{bmatrix}\dot{\theta} & 0 & 0\end{bmatrix}^{T}.
\end{equation}

The target of interest (UGV) is represented by the origin of the non-inertial frame $N$. The position of the non-inertial frame origin in the camera frame $C$ is:
\begin{equation}
    \label{p_C_N}
{}^C\B{p}_N = -\rotmat{C}{N}\,\nbx{\B{p}} + {}^C\B{p}_B,
\end{equation}

where ${}^C\B{p}_B$ denotes the position of the UAV body-frame origin expressed in the camera frame, which is determined by the gimbal mounting geometry and varies with the gimbal angle $\theta$.

Taking the time derivative of ${}^C\B{p}_N$:
\begin{equation}
\begin{aligned}
    \label{p_C_N_dot_raw}
{}^C\dot{\B{p}}_N &= -\dot{\rotmat{C}{N}}\,\nbx{\B{p}} - \rotmat{C}{N}\,\nbx{\B{v}} + {}^C\dot{\B{p}}_B.
\end{aligned}
\end{equation}

We assume that the camera optical center lies on the gimbal rotation axis. Since ${}^B\B{p}_C$ is a constant vector in frame $B$ fixed by the gimbal mounting geometry, applying Eq.~\eqref{R_C_B_DOT} to differentiate ${}^C\B{p}_B = -\rotmat{C}{B}\,{}^B\B{p}_C$ yields:
\begin{equation}
{}^C\dot{\B{p}}_B = -[{}^C\B{\omega}_C^B]_{\times}\,{}^C\B{p}_B.
\end{equation}

The rotation matrix $\rotmat{C}{N}$ can be decomposed as:
\begin{equation}
\rotmat{C}{N} = \rotmat{C}{B}\,\rotmat{B}{N},
\end{equation}
where $\rotmat{C}{B}$ is the gimbal rotation matrix depending on the gimbal angle $\theta$. The relative mounting configuration between the gimbal and the UAV body, along with the positive rotation direction of the gimbal, is illustrated in Fig.~\ref{relative}. For a single-axis gimbal rotating about the y-axis of the UAV body frame, the rotation matrix $\rotmat{C}{B}$ is given by:
\begin{equation}
\rotmat{C}{B} = \begin{bmatrix}
    0 & -1 & 0 \\
-\cos\theta & 0 & -\sin\theta \\
\sin\theta & 0 & -\cos\theta
\end{bmatrix}.
\end{equation}
This corresponds to a pitch rotation that allows the camera to tilt up and down relative to the UAV body frame. The gimbal's roll and yaw remain aligned with the body frame.

The derivative of the rotation matrix $\rotmat{C}{N}$ can be derived as:
\begin{equation}
\begin{aligned}
    \label{R_C_N_DOT}
\dot{\rotmat{C}{N}} &= \frac{d}{dt}\left(\rotmat{C}{B}\,\rotmat{B}{N}\right) \\[0.8em]
&= -\left([{}^C\B{\omega}_C^B]_{\times} + [\rotmat{C}{B}\,{}^B\B{\Omega}_B]_{\times}\right)\,\rotmat{C}{N} \\
&\quad + \rotmat{C}{N}\,\left[{}^N\B{\Omega}_N\right]_{\times}.
\end{aligned}
\end{equation}

Substituting the expression for $\dot{\rotmat{C}{N}}$ from Eq.~\eqref{R_C_N_DOT} into Eq.~\eqref{p_C_N_dot_raw}, we obtain:
\begin{equation}
    \label{p_C_N_DOT}
\begin{aligned}
{}^C\dot{\B{p}}_N
&= \left([{}^C\B{\omega}_C^B]_{\times} + [\rotmat{C}{B}\,{}^B\B{\Omega}_B]_{\times}\right)\,\rotmat{C}{N}\,\nbx{\B{p}} \\[0.4em]
&\quad - \rotmat{C}{N} \left[{}^N\B{\Omega}_N\right]_{\times}\,\nbx{\B{p}} - \rotmat{C}{N}\nbx{\B{v}} \\[0.4em]
&\quad - [{}^C\B{\omega}_C^B]_{\times}\,{}^C\B{p}_B.
\end{aligned}
\end{equation}

The four terms correspond to the gimbal-body composite rotation, the non-inertial effect of the UGV's rotating frame, the relative velocity, and the velocity induced by the camera-body offset.

With ${}^C\B{p}_N$ and ${}^C\dot{\B{p}}_N$ established, we define how the target projects onto the image plane. Let $I$ denote the projection plane. The projection is given by the pinhole camera model:
\begin{equation}
    \label{eq:projection}
{}^I\B{p}_{N} = \begin{bmatrix} u_I \\ v_I \end{bmatrix} = \begin{bmatrix} f_x \frac{{}^Cp_{N,x}}{{}^Cp_{N,z}} \\ f_y \frac{{}^Cp_{N,y}}{{}^Cp_{N,z}} \end{bmatrix},
\end{equation}
where $f_x$ and $f_y$ are the focal lengths, and $(u_I,v_I)$ are measured relative to the camera principal point.

Taking the time derivative of Eq.~\eqref{eq:projection}:
\begin{equation}
{}^I\dot{\B{p}}_{N} = \begin{bmatrix} f_x \frac{{}^C\dot{p}_{N,x} \, {}^Cp_{N,z} - {}^Cp_{N,x} \, {}^C\dot{p}_{N,z}}{({}^Cp_{N,z})^2} \\[1.5ex] f_y \frac{{}^C\dot{p}_{N,y} \, {}^Cp_{N,z} - {}^Cp_{N,y} \, {}^C\dot{p}_{N,z}}{({}^Cp_{N,z})^2} \end{bmatrix}.
\end{equation}
This image-plane velocity is used as the perception dynamics in the MPC.

Equivalently, the image-plane velocity can be written in the following compact form:
\begin{equation}
    \label{s_dot}
\begin{bmatrix} {}^I\dot{\B{p}}_N \\ 0 \end{bmatrix} =
\dfrac{1}{({}^Cp_{N,z})^2}
\begin{bmatrix}
0 & -f_x & 0 \\[0.8em]
f_y & 0 & 0 \\[0.8em]
0 & 0 & 0
\end{bmatrix}
\bigl({}^C\dot{\B{p}}_N \times {}^C\B{p}_N\bigr).
\end{equation}
This quantity ${}^I\dot{\B{p}}_N$ serves as the perception dynamics within the MPC. By jointly penalizing ${}^I\B{p}_N$, the controller drives the target toward the image center.

\subsection{Learned Target-Motion Prediction}
\label{sec:target_motion}

The active-perception dynamics model treats $\B{\Psi}$ as target-motion parameters, but its future values $\B{\Psi}_k$ for $k \in [1, H-1]$ are unknown at decision time. Here, $H$ denotes the number of steps in the MPC prediction horizon.
The baseline CoNi-MPC~\cite{zhang2023coni} uses a zero-order hold (ZOH) across the prediction steps, filling the horizon by repeating $\B{\Psi}_0$.
This approximation is reasonable during steady motion but loses accuracy when the UGV acceleration or angular velocity changes rapidly.
We replace the constant-parameter assumption with a learned Target-Motion Predictor that maps a short window of UGV motion history to $\B{\Psi}_{0:H-1}$.

For a differential-drive UGV operating on flat terrain, the longitudinal acceleration $a_x$ and yaw rate $\omega_z$ are selected as motion primitives. The network predicts these two channels from the target-motion history, producing $\hat{\B{y}}_{0:H-1}\in\mathbb{R}^{H\times2}$. They are then converted into the target-motion parameters required by the non-inertial dynamics, where ${}^N\B{a}_N=[a_x,\ v_x\omega_z,\ 0]^T$ and ${}^N\B{\Omega}_N=[0,\ 0,\ \omega_z]^T$. Here, $v_x$ is propagated from the current longitudinal velocity using the predicted acceleration.

\subsubsection{Network Architecture}

We use a TCN~\cite{bai2018empirical} as the backbone. 
Compared with a recurrent model, it trains in parallel across time and has deterministic inference latency.
The backbone stacks dilated causal 1D convolutions, so each output step sees a long history.
An MLP head then maps the final temporal feature to the prediction horizon.
Instead of the absolute sequence, the head outputs a residual:
\begin{equation}
    \hat{\B{y}}_k = \B{y}_{\mathrm{latest}} + \B{\delta}_k, \qquad k = 0, \dots, H-1,
    \label{eq:residual_pred}
\end{equation}
where $\B{y}_{\mathrm{latest}}$ is the most recent target-motion observation broadcast across the horizon and $\B{\delta}_k$ is the network correction.
Initializing the final layer's bias to zero makes the predictor reproduce the constant-parameter baseline at the start of training.
The network then only has to learn the correction term.

\subsubsection{Loss Function}

A uniform MSE weights every horizon step equally, yet parameter errors at different steps affect the predicted state to different degrees.
For the discrete dynamics $\B{x}_{k+1} = f_{\B{\Psi}_k}(\B{x}_k, \B{u}_k)$ under a fixed control sequence, let $L_x$ and $L_{\B{\Psi}}$ be the local Lipschitz constants of $f$ with respect to the state and the parameter over the prediction region.
A parameter error $\delta\B{\Psi}_k$ propagates through the dynamics. 
Summed over the horizon, the terminal state error $\delta\B{x}_{H}$ satisfies
\begin{equation}
    \|\delta \B{x}_{H}\| \;\le\; \sum_{k=0}^{H-1} L_x^{\,H-1-k}\, L_{\B{\Psi}}\, \|\delta \B{\Psi}_k\|,
    \label{eq:lipschitz_bound}
\end{equation}
so an error at step $k$ is amplified by $L_x^{H-1-k}$ before it reaches the horizon end. 
The earliest steps carry the largest factor.
Since $L_x^{H-1-k} \propto (1/L_x)^k$, matching this profile calls for weights that decay geometrically in $k$:
\begin{equation}
    \mathcal{L}_{\text{wMSE}} = \sum_{k=0}^{H-1} w_k \big\|\hat{\B{y}}_k - \B{y}_k\big\|^2,
    \quad w_k = \frac{\gamma^k}{\sum_{j=0}^{H-1} \gamma^j}.
    \label{eq:wmse}
\end{equation}
Here $\gamma \in (0,1)$ sets the decay rate, and the bound suggests $\gamma \approx 1/L_x$.
To suppress chattering predictions that would induce oscillations in the MPC, we add a first-order finite-difference penalty:
\begin{equation}
    \mathcal{L}_{\text{smooth}} = \frac{1}{H-1} \sum_{k=1}^{H-1} \big\|\hat{\B{y}}_k - \hat{\B{y}}_{k-1}\big\|^2.
    \label{eq:smooth}
\end{equation}
The total objective is $\mathcal{L} = \mathcal{L}_{\text{wMSE}} + \lambda_s \mathcal{L}_{\text{smooth}}$, where $\lambda_s$ weights the smoothness term.

\subsubsection{Training Data}

We record UGV motion under randomized command profiles, covering a wide range of acceleration and turning regimes.
The two-channel signal is filtered with a low-pass filter to suppress sensor noise without adding lag.
It is then sliced into sliding windows and split chronologically into training, validation, and test sets. 
A single standardization, fit on the training targets, is applied to both inputs and outputs. 
This keeps the anchor $\B{y}_{\mathrm{latest}}$ and the network correction $\B{\delta}_k$ in Eq.~\eqref{eq:residual_pred} in the same standardized space.
Online augmentation applies per-channel amplitude scaling to both the input histories and target sequences, and Gaussian noise to the input histories only.

\subsection{Optimization Problem Formulation}
\label{sec:mpc}
We now incorporate the active-perception dynamics model of Section~\ref{sec:dynamics} and the predicted target motion of Section~\ref{sec:target_motion} into a unified MPC formulation.
Combining Eq.~\eqref{eq:relative_dynamics} with the perception-related dynamics from Eq.~\eqref{p_C_N}, \eqref{p_C_N_DOT}, and \eqref{s_dot}, we obtain the complete active-perception dynamics for the system state and control vector defined above.

The active-perception dynamics are written compactly as:
\begin{equation}
    \dot{\B{x}} = f_{\B{\Psi}}(\B{x}, \B{u}),
    \label{eq:extended_dynamics}
\end{equation}
where $\B{\Psi} = [\nnx{\B{a}}; \nnx{\B{\Omega}}]$ is the target-motion parameter vector supplied by the UGV.

The core of our framework is a nonlinear MPC that optimizes $\B{u}$ over the $H$-step prediction horizon. 
The cost function jointly penalizes relative trajectory tracking error, image-plane deviation, control effort, and terminal state deviation:
\begin{equation}
\begin{aligned}
    \min_{\B{u}_0, \dots, \B{u}_{H-1}} \quad
    & \sum_{k=0}^{H-1} ( \|\B{x}_k - \B{x}_{\mathrm{ref},k}\|_{\B{Q}}^2 + \|\B{u}_k\|_{\B{R}}^2 ) \\
    & + \|\B{x}_{H} - \B{x}_{\mathrm{ref},H}\|_{\B{Q}_{\mathrm{final}}}^2.
\end{aligned}
\label{eq:mpc_cost}
\end{equation}
For the image-plane components of $\B{x}_{\mathrm{ref},k}$, the reference is set to the image center.

The optimization is subject to the dynamics, actuator bounds, and FOV constraints for all predictions $k \in [0, H-1]$:
\begingroup
\interdisplaylinepenalty=0

\begin{subequations}
\label{eq:mpc_constraints}
\begin{IEEEeqnarray}{rCl}
    \IEEEeqnarraymulticol{3}{c}{\B{x}_0 = \B{x}(t_0),}
    \label{eq:mpc_initial}\\
    \IEEEeqnarraymulticol{3}{c}{
        \B{x}_{k+1}=f_{\B{\Psi}_k}(\B{x}_k,\B{u}_k),}
    \label{discrete}\\
    \IEEEeqnarraymulticol{3}{c}{
        T_{\min}\leq T_k\leq T_{\max},}
    \label{eq:mpc_thrust}\\
    \IEEEeqnarraymulticol{3}{c}{
        \left|[{}^B\B{\Omega}_{B,k}]_i\right|
        \leq\Omega_{B,\max},\quad i\in\{x,y,z\},}
    \label{eq:mpc_rate}\\
    \IEEEeqnarraymulticol{3}{c}{
        \theta_{\min}\leq\theta_k\leq\theta_{\max},
        \quad|\dot{\theta}_k|\leq\dot{\theta}_{\max},}
    \label{eq:mpc_gimbal_bounds}\\
    \IEEEeqnarraymulticol{3}{c}{
        |u_{I,k}|\leq u_{I,\max},
        \quad|v_{I,k}|\leq v_{I,\max},}
    \label{eq:mpc_fov}\\
    \IEEEeqnarraymulticol{3}{c}{
        {}^Cp_{N,z,k}\geq z_{\min}>0,}
    \label{eq:Z}
\end{IEEEeqnarray}
\end{subequations}

\endgroup

where $\B{x}_k$ and $\B{u}_k$ are the state vector and the control vector at step $k$; $\B{x}_{\mathrm{ref},k}$ is the corresponding reference trajectory; and $\B{Q}$, $\B{R}$, $\B{Q}_{\mathrm{final}}$ are positive definite weighting matrices for state tracking, control effort, and terminal cost, respectively. The image-plane limits $u_{I,\max}$ and $v_{I,\max}$ are determined by the camera FOV. In Eq.~\eqref{eq:Z}, ${}^Cp_{N,z,k}$ is the target depth along the camera optical axis. Its positive lower bound keeps the target in front of the camera. Constraints~\eqref{eq:mpc_initial}--\eqref{eq:Z} enforce the system dynamics and initial condition, respect the actuator limits, and keep the target within the FOV.

At each control step, the MPC tracks the desired relative trajectory while adjusting the gimbal to maintain target visibility. The target-motion parameters $\B{\Psi}_k$ in Eq.~\eqref{discrete} are supplied by the Target-Motion Predictor of Section~\ref{sec:target_motion} as the horizon-length sequence $\hat{\B{\Psi}}_{0:H-1}$.

\subsection{Implementation Details}

\subsubsection{Target-Motion Predictor}

\begin{table}[t]
    \centering
    \caption{Target-Motion Predictor Hyperparameters}
    \label{tab:tcn_hyper}
    \small
    \renewcommand{\arraystretch}{1.15}
    \begin{tabular}{l l}
        \toprule
        \multicolumn{2}{c}{\textbf{Network}} \\
        \midrule
        Input window & $25 \times 2$ at $50\,\mathrm{Hz}$ \\
        Output horizon & $20 \times 2$ at 10\,Hz \\
        Residual blocks & 4, channels $[128, 192, 256, 320]$ \\
        Kernel / dilations & 3 / $[1, 2, 4, 8]$ \\
        Dropout & 0.2 \\
        Parameters & $\approx 1.35\times10^6$ \\
        \midrule
        \multicolumn{2}{c}{\textbf{Loss}} \\
        \midrule
        Weights ($\gamma$, $\lambda_s$) & 0.9, 0.01 \\
        \midrule
        \multicolumn{2}{c}{\textbf{Training}} \\
        \midrule
        Dataset & 60\,min, $\sim184{,}000$ frames \\
        Butterworth cutoff & 5\,Hz \\
        Train/val/test split & 80\%/15\%/5\% \\
        Optimizer & AdamW, weight decay $10^{-3}$ \\
        LR schedule & cosine, $5{\times}10^{-4} \to 10^{-6}$ \\
        Batch / epochs & 256 / 500 (patience 60) \\
        \bottomrule
    \end{tabular}
\end{table}

Table~\ref{tab:tcn_hyper} lists the network and training configuration. 
The residual blocks follow the standard TCN design~\cite{bai2018empirical}, with a receptive field that spans the full input window.

At deployment, the predictor maintains a 25-frame circular target-motion buffer sampled at $50\,\mathrm{Hz}$ and runs inference at $50\,\mathrm{Hz}$. Each inference produces 20 prediction points at 0.1\,s intervals, covering the next 2\,s.
It inverse-transforms the output to physical units and clips each channel to its training range, widened by 20\% on each side. 
The latest prediction sequence is made available to the MPC between predictor updates.

\subsubsection{MPC Solver}
The MPC control loop runs stably at approximately 110\,Hz. It uses a horizon of $H = 20$ steps, giving a 2\,s look-ahead with a prediction step of $\Delta t = 0.1$\,s. The continuous dynamics of Eq.~\eqref{eq:extended_dynamics} are discretized with RK4.
The optimization problem is solved with acados~\cite{Verschueren2021}, using the SQP Real-Time Iteration (RTI) scheme with Gauss-Newton Hessian approximation. The QP subproblems are solved with the partial-condensing HPIPM solver and a condensing horizon of $N_{\text{cond}}=5$. The formulation uses a maximum of 50 QP iterations and a feasibility tolerance of $10^{-6}$.

\section{Experiments}

We evaluate COPA through simulation comparisons, predictor ablations, and real-world tracking experiments.

\subsection{Comparison Experiments}

All comparative experiments were conducted in Gazebo using the PX4 Software-in-the-Loop (SITL) framework to improve simulation fidelity. The maximum linear velocities of the UAV and UGV were set to 12\,m/s and 10\,m/s, respectively, while their maximum accelerations were both limited to $3\,\mathrm{m/s^2}$. The maximum UAV body rate and UGV yaw rate were both set to $1.0\,\mathrm{rad/s}$. The simulated camera had a resolution of $640\times480$ pixels and horizontal and vertical FOVs of $106^\circ$ and $93^\circ$, respectively.
For evaluation, we generate 30 randomized UGV-motion scenarios shared across all methods. For trial \(j\), the same random seed, UGV trajectory, and initial condition are used for every method. Each trajectory lasted 120\,s, with new linear-velocity and yaw-rate commands sampled every 10\,s.

We compare the proposed method against the following three baseline methods:

\begin{itemize}
    \item \textbf{CoNi} \cite{zhang2023coni}: Original CoNi-MPC augmented with a body-fixed camera; no active perception or FOV constraints.
    \item \textbf{CoNi-POI}: A point-of-interest perception cost adapted from PAMPC \cite{falanga2018pampc} is integrated into the CoNi non-inertial MPC, with the camera fixed at $0^\circ$; this baseline isolates perception-aware optimization from gimbal control.
    \item \textbf{CoNi-PID}: CoNi for UAV flight with an independent visual-servoing PID gimbal. This baseline evaluates the limitations of decoupled control under dynamic maneuvers.
\end{itemize}

For a fair comparison, all methods hold the latest target-motion parameters constant over the prediction horizon. This setting allows us to assess the contributions of the active gimbal and the unified MPC independently of target-motion prediction.

The objective function is a composite loss defined as:
\begin{equation}
\text{Loss} = \text{RMSE} + w_{\mathrm{uv}} \cdot \frac{\bar{d}_{\mathrm{uv}}}{d_{\mathrm{norm}}} + w_{\mathrm{fov}} \cdot r_{\mathrm{fov}} + w_{\mathrm{fail}} \cdot \delta_{\mathrm{fail}},
\end{equation}
where $\bar{d}_{\mathrm{uv}}$ is the mean pixel distance of the target from the image center, $d_{\mathrm{norm}} = 320$\,px is a scale factor that brings $\bar{d}_{\mathrm{uv}}$ to the same magnitude as the other loss terms. $r_{\mathrm{fov}} \in [0,1]$ is the fraction of frames in which the target leaves the FOV and $\delta_{fail}\in\{0,1\}$ is a binary indicator of tracking failure. The weights are set to $w_{\mathrm{fail}} = 50$, $w_{\mathrm{uv}} = 5$, and $w_{\mathrm{fov}} = 1$, reflecting the priority order: tracking loss $\gg$ image centering $>$ RMSE $\approx$ FOV violation.

\begin{figure}[htbp]
    \centering
    \includegraphics[width=\columnwidth]{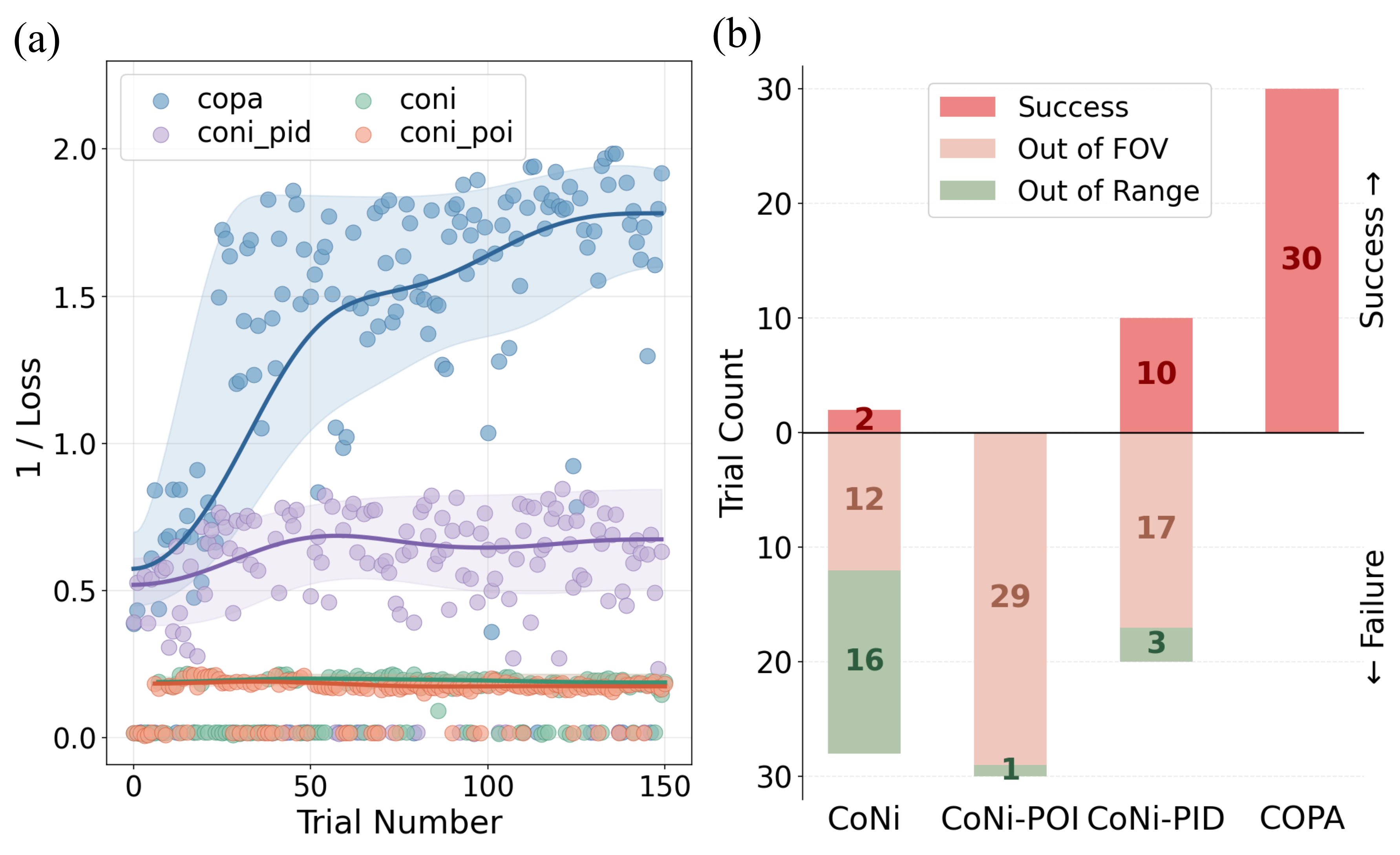}
    \caption{(a) Hyperparameter optimization convergence over 150 trials per method (reciprocal loss; higher is better). (b) Trial outcome breakdown across 30 runs per method, showing counts of successful trials, Out-of-FOV failures, and Out-of-Range failures.}
    \label{fig:optuna_three_rate}
\end{figure}

Each method uses a dedicated 150-trial Optuna study~\cite{akiba2019optuna} with the Tree-structured Parzen Estimator algorithm. The tunable parameters are method-specific. CoNi optimizes the cost weights $\B{Q}$ and $\B{R}$ of its tracking controller. CoNi-POI additionally optimizes the image-plane weights $Q_{u_I}$ and $Q_{v_I}$. CoNi-PID jointly tunes the tracking cost weights and the gimbal PID gains ($K_p$, $K_i$, $K_d$). COPA optimizes $\B{Q}$, $\B{R}$, $Q_{u_I}$, $Q_{v_I}$, $Q_\theta$, and $R_{\dot{\theta}}$ within the proposed unified MPC formulation. The convergence histories are shown in Fig.~\ref{fig:optuna_three_rate}(a).

\subsubsection{Performance Under Varied Target Dynamics}
\label{sec:exp1}
In the first set of experiments, we assess robustness to dynamically varying target motions within the tested operating range.
The UAV executes relative circular tracking while the UGV moves randomly.
We compare tracking error (RMSE), target visibility rate, and mission success rate among the four methods.

We distinguish two failure modes. 
An Out-of-Range failure occurs when the horizontal UAV--UGV distance exceeds 5\,m, corresponding to the maximum reliable perception range of the real-world system.
An Out-of-FOV failure occurs when the target temporarily leaves the camera FOV while remaining within the valid perception range. Fig.~\ref{fig:optuna_three_rate}(b) shows the breakdown across 30 runs per method. COPA completes 30 of 30 trials successfully. CoNi fails almost entirely due to FOV loss. CoNi-PID improves FOV retention but yields only 10 successes.

\begin{figure}[htbp]
    \centering
    \includegraphics[width=\columnwidth]{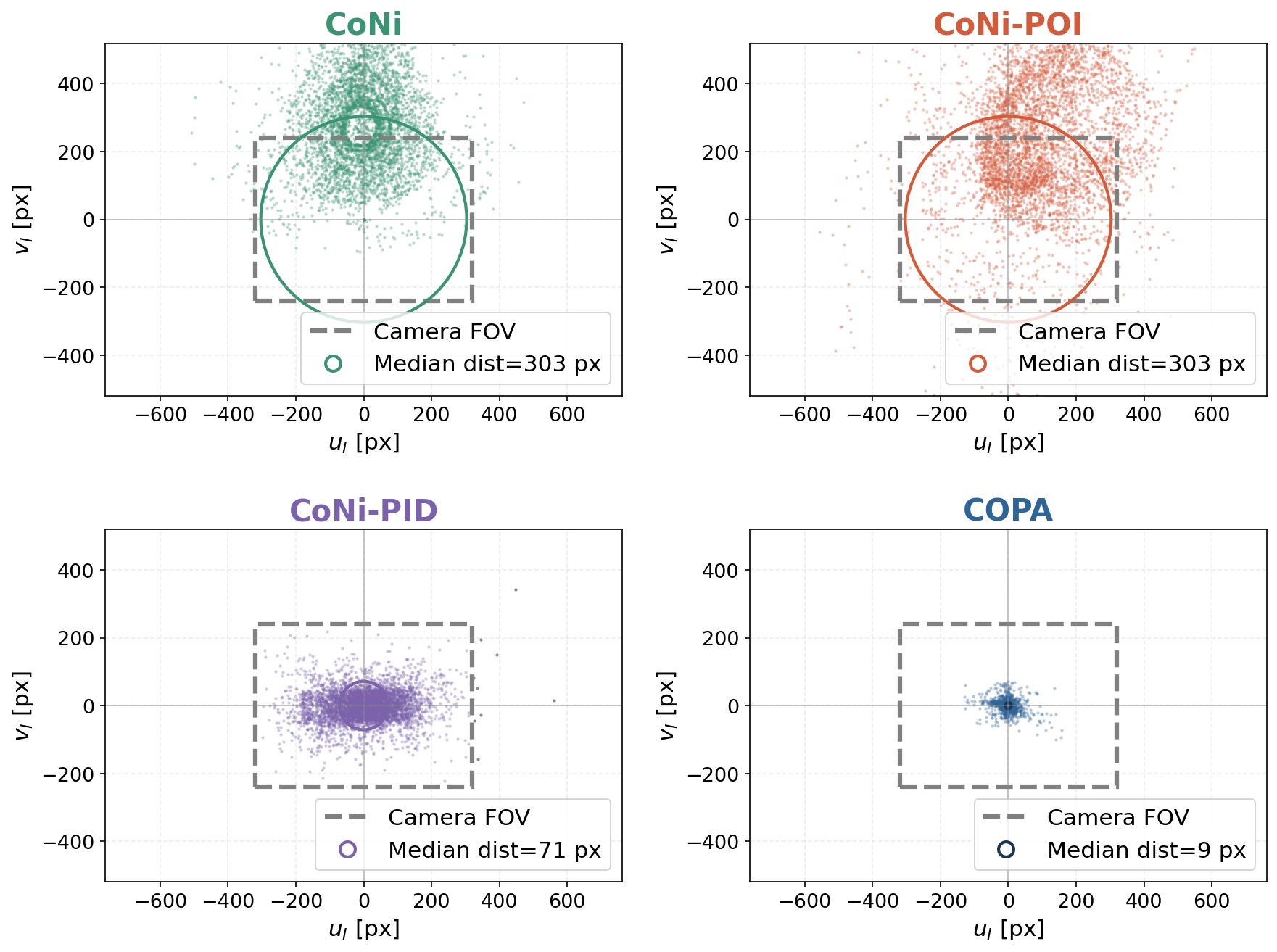}
    \caption{Target image-plane positions $(u_I,v_I)$ across 30 randomized UGV trajectories under circular tracking. Each point represents one frame; the dashed rectangle denotes the camera FOV, and each circle reports the median distance to the image center.}
    \label{fig:point_scatter}
\end{figure}

Fig.~\ref{fig:point_scatter} shows the distribution of target image coordinates $(u_I, v_I)$. Each point is one frame. CoNi and CoNi-POI both scatter near the FOV boundary (303\,px median). CoNi-POI's image-plane cost biases points toward the center, but centering the target requires the UAV to tilt, which conflicts with trajectory tracking. CoNi-PID's reactive gimbal reduces the median to 71\,px. However, since the body MPC does not include the image-plane state, the UAV makes no proactive effort to keep the target in view. COPA achieves 9\,px by co-optimizing the gimbal and trajectory in a unified MPC.

\begin{figure}[htbp]
    \centering
    \includegraphics[width=.92\columnwidth]{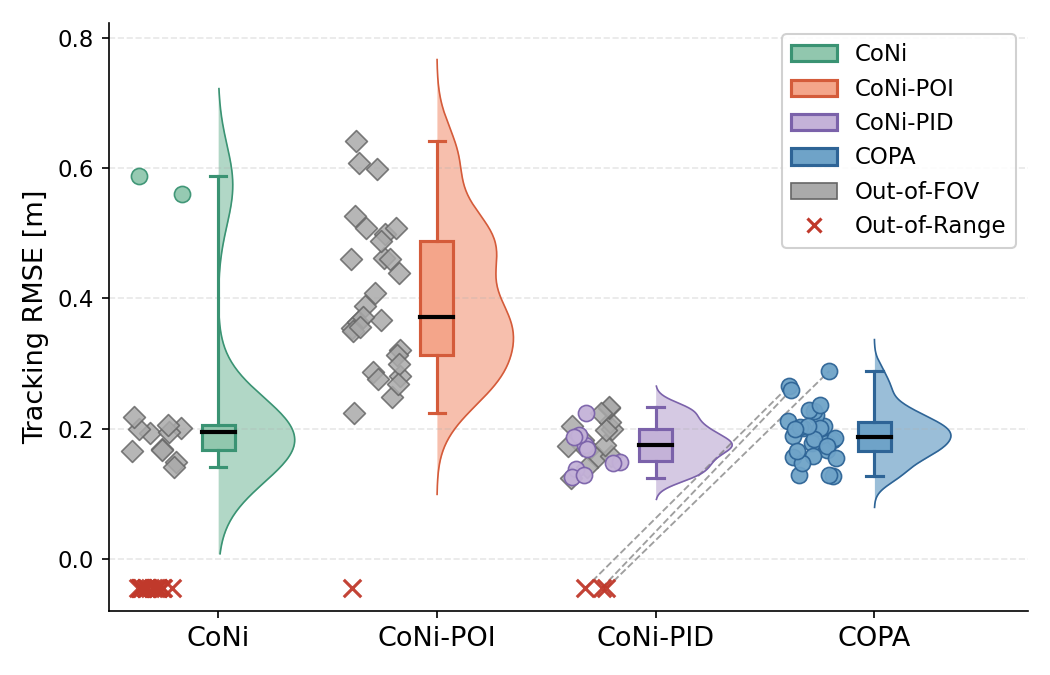}
    \caption{Raincloud plot of tracking RMSE across four methods. The gray dotted lines connect the three paired trials in which CoNi-PID incurs an Out-of-Range failure while COPA completes the task successfully.}
    \label{fig:rmse_success}
\end{figure}

Fig.~\ref{fig:rmse_success} includes Out-of-FOV trials in the RMSE evaluation, as the simulator provides the ground-truth relative state even when the target temporarily leaves the camera FOV. Out-of-Range trials are excluded because the target exceeds the reliable perception range and the corresponding tracking error is no longer meaningful.
CoNi succeeds only on two trajectories where the target stays in view. CoNi-POI completes no successful trials, further confirming that a perception-aware MPC without gimbal decoupling cannot resolve the conflict between body attitude and target visibility. Compared with CoNi-PID, COPA completes three additional trials, succeeding in all 30 while maintaining a low RMSE and incurring no observed Out-of-FOV failures. This demonstrates robustness to randomized target-motion variations.

\subsubsection{Performance Boundary Under Scaled Task Complexity}
\label{sec:exp2}

\begin{figure}[htbp]
    \centering
    \includegraphics[width=.92\columnwidth]{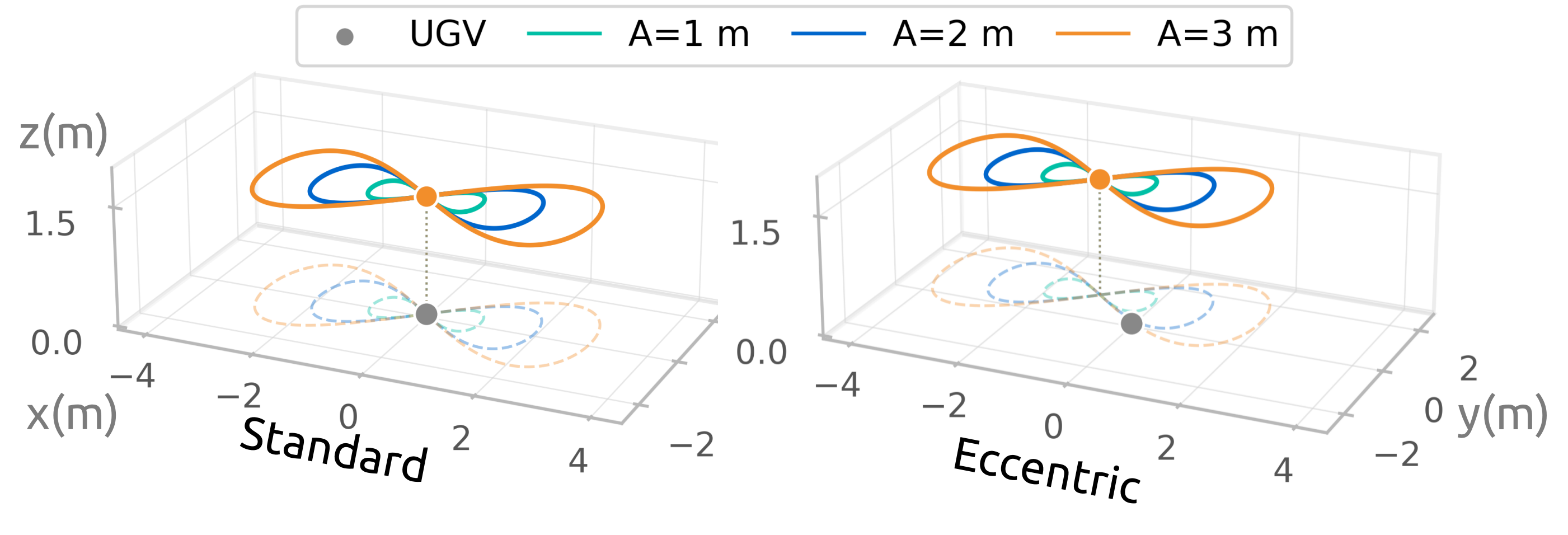}
    \caption{Standard and eccentric Figure-8 relative trajectories with three spatial scales ($A=1$, $2$, and $3$\,m).}
    \label{fig:8shape}
\end{figure}

In the second set of experiments, we fix the UGV motion and vary the spatial scale of the UAV relative trajectories to isolate the effect of increasing task demands. The Standard Figure-8 and Eccentric Figure-8 maneuvers shown in Fig.~\ref{fig:8shape} are tested at three scales, where Small, Medium, and Large correspond to trajectory amplitudes of $A=1$, $2$, and $3$\,m, respectively. Larger amplitudes and asymmetric trajectories progressively stress relative tracking and target visibility. The tracking RMSE, visibility rate, and mean image distance are summarized in Table~\ref{tab:quantitative_results}.

As the scale increases, CoNi suffers from degraded visibility due to the coupling between aggressive UAV motion and the body-fixed camera. It maintains partial tracking at medium scale in the Standard Figure-8 but loses visibility in larger-scale or asymmetric maneuvers. CoNi-POI completes all scenarios but exhibits larger RMSE (1.07--1.23\,m) and increased image deviations, while CoNi-PID improves performance in Standard Figure-8 but fails under medium and large Eccentric Figure-8 cases. In contrast, COPA achieves 100\% visibility, RMSE below 0.5\,m, and mean image distances ranging from 25.4 to 32.8\,px across all scenarios, demonstrating improved robustness under large-scale and asymmetric relative motions.

\begin{table*}[t]
    \centering
    \small
    \caption{Quantitative performance comparison across various trajectory types and scales. Arrows $\uparrow$ and $\downarrow$ indicate that higher and lower values are better, respectively. A dash (---) indicates an Out-of-Range failure.}
    \label{tab:quantitative_results}
    \renewcommand{\arraystretch}{1.2}
    
    \setlength{\tabcolsep}{2pt}
    \resizebox{\textwidth}{!}{
    \begin{tabular}{c c cccc c cccc c cccc}
        \toprule
        
        \multicolumn{2}{c}{\multirow{2}{*}{\textbf{Scenario}}} 
        & \multicolumn{4}{c}{\textbf{Tracking RMSE (m) $\downarrow$}} 
        & & \multicolumn{4}{c}{\textbf{Visibility Rate (\%) $\uparrow$}} 
        & & \multicolumn{4}{c}{\textbf{Mean Image Distance (px) $\downarrow$}} \\

        \cmidrule(lr){3-6} \cmidrule(lr){8-11} \cmidrule(l){13-16}

        & & \textbf{\small CoNi} & \textbf{\small CoNi-POI} & \textbf{\small CoNi-PID} & \textbf{\small COPA}
        & & \textbf{\footnotesize CoNi} & \textbf{\small CoNi-POI} & \textbf{\small CoNi-PID} & \textbf{\small COPA}
        & & \textbf{\footnotesize CoNi} & \textbf{\small CoNi-POI} & \textbf{\small CoNi-PID} & \textbf{\small COPA} \\
        
        \midrule
        
        \multirow{3}{*}{\makecell[c]{Standard \\ Figure-8}} 
        & Small  & 0.65 & 1.07 & 0.34 & \cellcolor{bestgray}\textbf{0.33} & & 97.1 & 97.1 & 99.1 & \cellcolor{bestgray}\textbf{100.0} & & 144.0{\scriptsize{$\pm$85.2}} & 143.7{\scriptsize{$\pm$82.4}} & 63.1{\scriptsize{$\pm$66.5}} & \cellcolor{bestgray}\textbf{29.9{\scriptsize{$\pm$31.8}}} \\
        & Medium & 0.65 & 1.12 & 0.38 & \cellcolor{bestgray}\textbf{0.37} & & 83.0 & 92.0 & 97.9 & \cellcolor{bestgray}\textbf{100.0} & & 209.8{\scriptsize{$\pm$124.5}} & 199.8{\scriptsize{$\pm$116.5}} & 63.0{\scriptsize{$\pm$64.7}} & \cellcolor{bestgray}\textbf{27.3{\scriptsize{$\pm$26.6}}} \\
        & Large  & ---  & 1.23 & \cellcolor{bestgray}\textbf{0.39} & \cellcolor{bestgray}\textbf{0.39} & & ---  & 72.1 & 97.5 & \cellcolor{bestgray}\textbf{100.0} & & --- & 268.3{\scriptsize{$\pm$153.1}} & 64.4{\scriptsize{$\pm$61.1}} & \cellcolor{bestgray}\textbf{31.4{\scriptsize{$\pm$36.4}}} \\

        \midrule

        \multirow{3}{*}{\makecell[c]{Eccentric \\ Figure-8}}
        & Small  & 0.66 & 1.08 & 0.36 & \cellcolor{bestgray}\textbf{0.35} & & 98.6 & 98.2 & 97.6 & \cellcolor{bestgray}\textbf{100.0} & & 125.6{\scriptsize{$\pm$70.7}} & 126.1{\scriptsize{$\pm$71.9}} & 81.0{\scriptsize{$\pm$71.8}} & \cellcolor{bestgray}\textbf{25.4{\scriptsize{$\pm$23.9}}} \\
        & Medium & ---  & 1.11 & ---  & \cellcolor{bestgray}\textbf{0.46} & & ---  & 94.4 & --- & \cellcolor{bestgray}\textbf{100.0} & & --- & 187.1{\scriptsize{$\pm$101.7}} & --- & \cellcolor{bestgray}\textbf{32.8{\scriptsize{$\pm$34.7}}} \\
        & Large  & ---  & 1.22 & ---  & \cellcolor{bestgray}\textbf{0.37} & & ---  & 80.3 & --- & \cellcolor{bestgray}\textbf{100.0} & & --- & 257.7{\scriptsize{$\pm$135.9}} & --- & \cellcolor{bestgray}\textbf{27.0{\scriptsize{$\pm$33.3}}} \\        
        \bottomrule
    \end{tabular}
    }
\end{table*}

\subsection{Ablation Study}
\label{sec:ablation}
We compare the ZOH baseline (w/o TCN) with horizon prediction (w/ TCN). Across 200 paired trials, TCN raises success from 92.0\% (184/200) to 97.5\% (195/200). Of 21 discordant pairs, 16 favor TCN and five favor ZOH; a continuity-corrected McNemar test gives $\chi^2(1)=4.762$, $p=0.029$. Mean RMSE changes only from 0.383 to 0.379\,m because model mismatch is concentrated in brief transients.
\begin{figure}[htbp]
    \centering
    \includegraphics[width=\columnwidth]{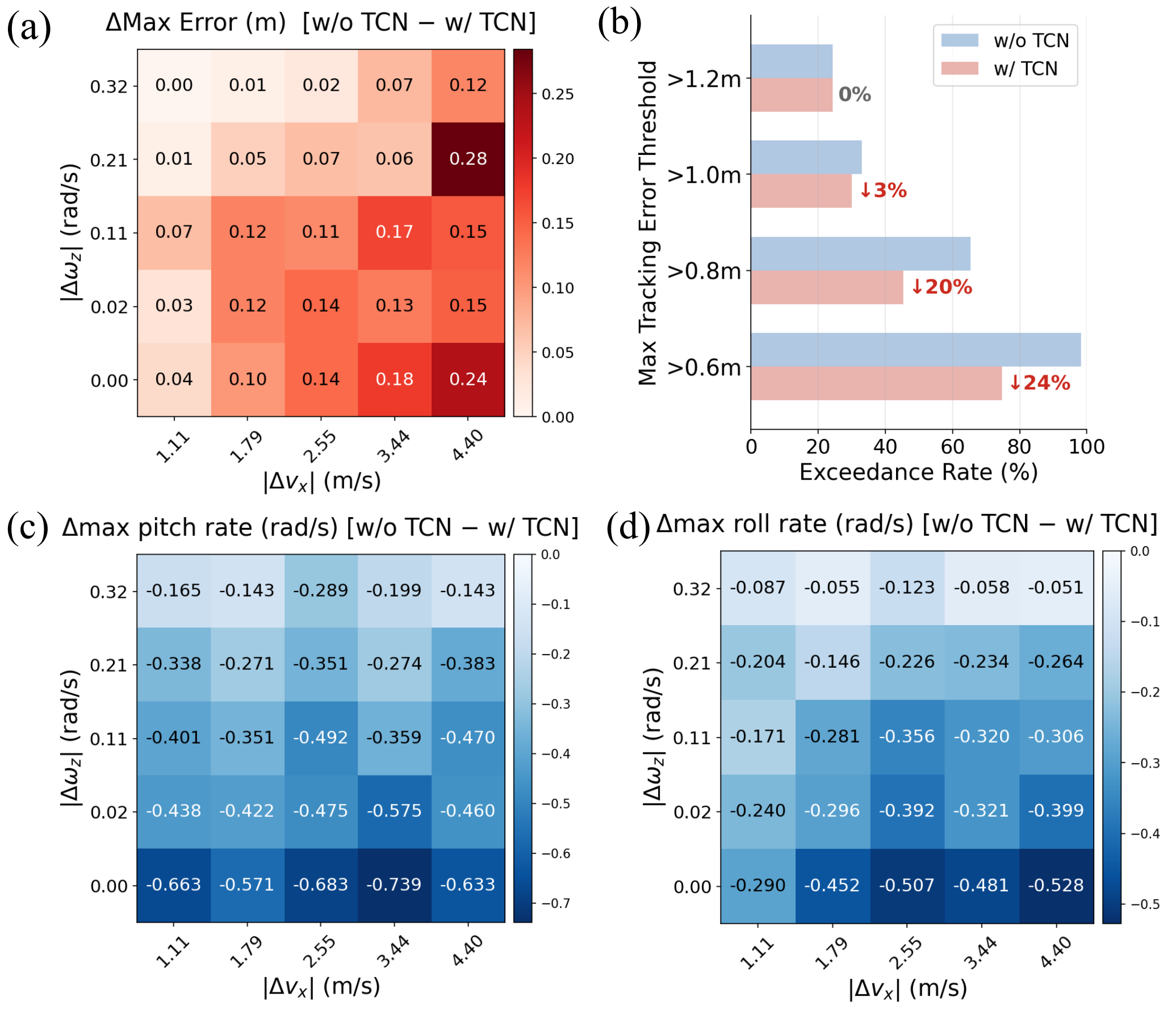}
    \caption{(a) Reduction in maximum tracking error (w/o TCN $-$ w/ TCN, m) across UGV motion change magnitudes. (b) Proportion of trials where maximum tracking error exceeds each threshold, with and without TCN. (c) and (d) Differences in peak pitch and roll rates within 2\,s of a motion change (w/o TCN $-$ w/ TCN, rad/s).}
    \label{fig:tcn_ablation}
\end{figure}

In Fig.~\ref{fig:tcn_ablation}, $|\Delta v_x|$ and $|\Delta\omega_z|$ denote commanded speed-step and yaw-rate changes. Because $|\Delta v_x|$ is directly controlled and correlates with the resulting longitudinal acceleration, it is used as a proxy for the longitudinal-transition magnitude. Panel~(a) shows positive paired maximum-error reductions in every bin, weakest at the largest $|\Delta\omega_z|$. Panel~(b) shows reductions of 24\% and 20\% in the exceedance rates at 0.6 and 0.8\,m, respectively, but little change above 1.0\,m. The negative pitch/roll-rate differences in (c) and (d) indicate a stronger TCN-enabled response, especially in pitch, consistent with reduced longitudinal model mismatch.

\begin{figure}[htbp]
    \centering
    \includegraphics[width=\columnwidth]{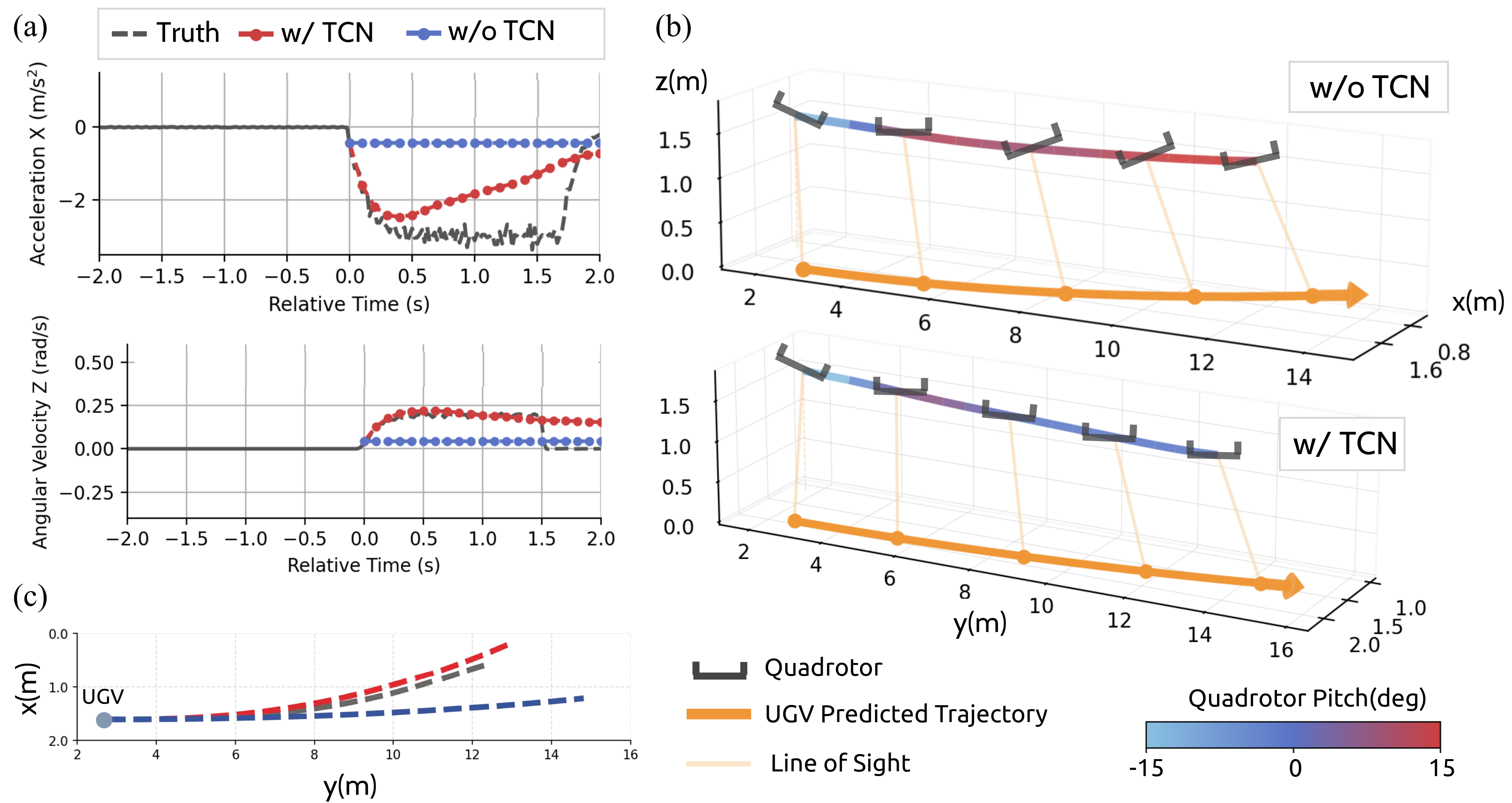}
    \caption{Relative time $t=0$ marks the onset of the UGV motion transition. (a) TCN (red) and constant-parameter (blue) predictions of UGV $a_x$ and $\omega_z$ vs.\ ground truth. (b) MPC-planned UAV pitch over the 2\,s horizon with and without TCN. Color encodes the pitch angle in degrees. (c) Predicted UGV future trajectories under the conditions in~(a).}
    \label{fig:tcn_mechanism}
\end{figure}

Fig.~\ref{fig:tcn_mechanism} illustrates a deceleration-and-turn transition beginning at $t=0$. As shown in Fig.~\ref{fig:tcn_mechanism}(a), the TCN predicts the subsequent evolution of $a_x$ and $\omega_z$, whereas the baseline holds their latest measurements constant. The resulting curved trajectory and pitch adjustment in Fig.~\ref{fig:tcn_mechanism}(c) and (b), respectively, illustrate a representative transition response consistent with the aggregate peak pitch rate and roll rate differences reported in Fig.~\ref{fig:tcn_ablation}(c) and (d).

\subsection{Real-World Experiments}

\begin{figure}[t]
    \centering
    \includegraphics[width=.90\columnwidth]{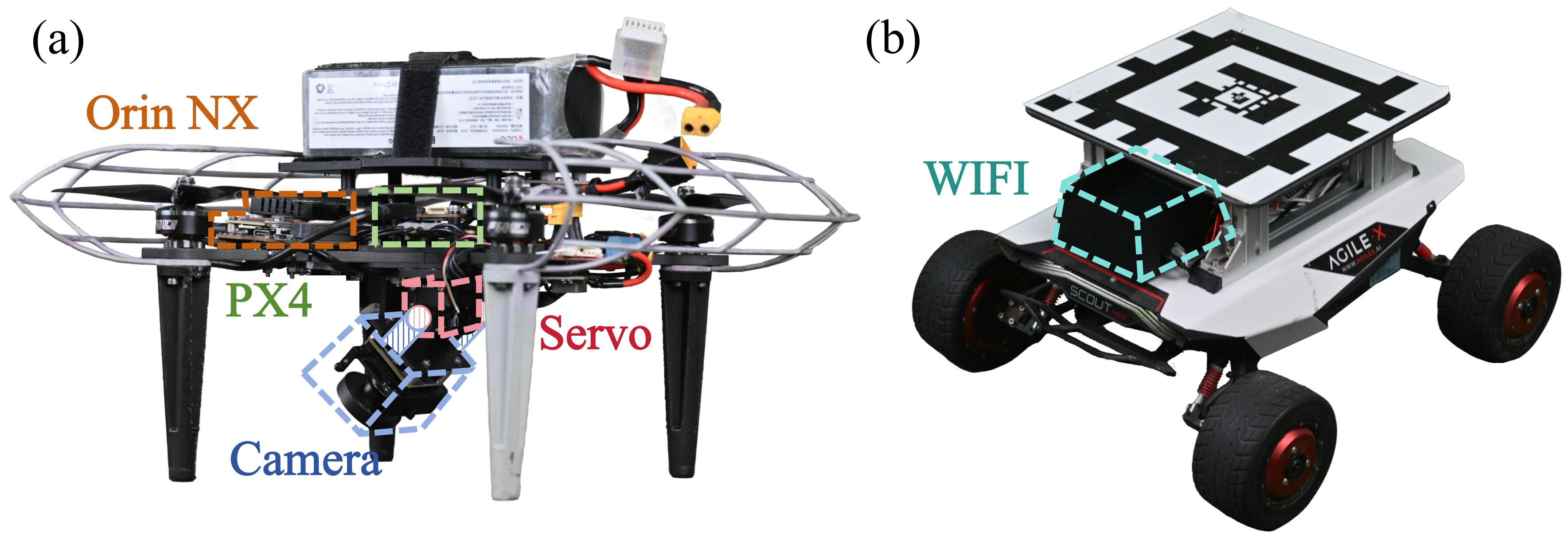}
    \caption{Real-world experimental platforms. (a) UAV platform. (b) UGV platform equipped with an AprilTag bundle.}
    \label{fig:platform}
\end{figure}

Our quadrotor platform (Fig.~\ref{fig:platform}), with a weight of 2\,kg, achieves a thrust-to-weight ratio of 2.5 when powered by a 6S battery.
The onboard computer is an NVIDIA Jetson Orin NX, and the flight controller is a Pixhawk 4. The perception system utilizes an MVSUA camera with a horizontal FOV of $106^\circ$ and a vertical FOV of $93^\circ$, mounted on a custom single-axis servo gimbal providing pitch rotation within $[-83^\circ, 83^\circ]$ and a maximum angular velocity of $180^\circ/\mathrm{s}$.

The UGV estimates its longitudinal acceleration and yaw rate from wheel-encoder measurements. These target-motion estimations are streamed to the UAV via Wi-Fi. The UGV carries a nested AprilTag bundle (Tag36h11 family) composed of three concentric tags with an approximately 5:1 side-length ratio which enables reliable visual detection from long-range to close-proximity operation. Indoor calibration against the NOKOV\footnote{\url{https://www.nokov.com/}} motion capture system yields an RMSE of 3.4\,cm, confirming sufficient accuracy for outdoor evaluation.

\begin{figure}[t]
    \centering
    \includegraphics[width=.96\columnwidth]{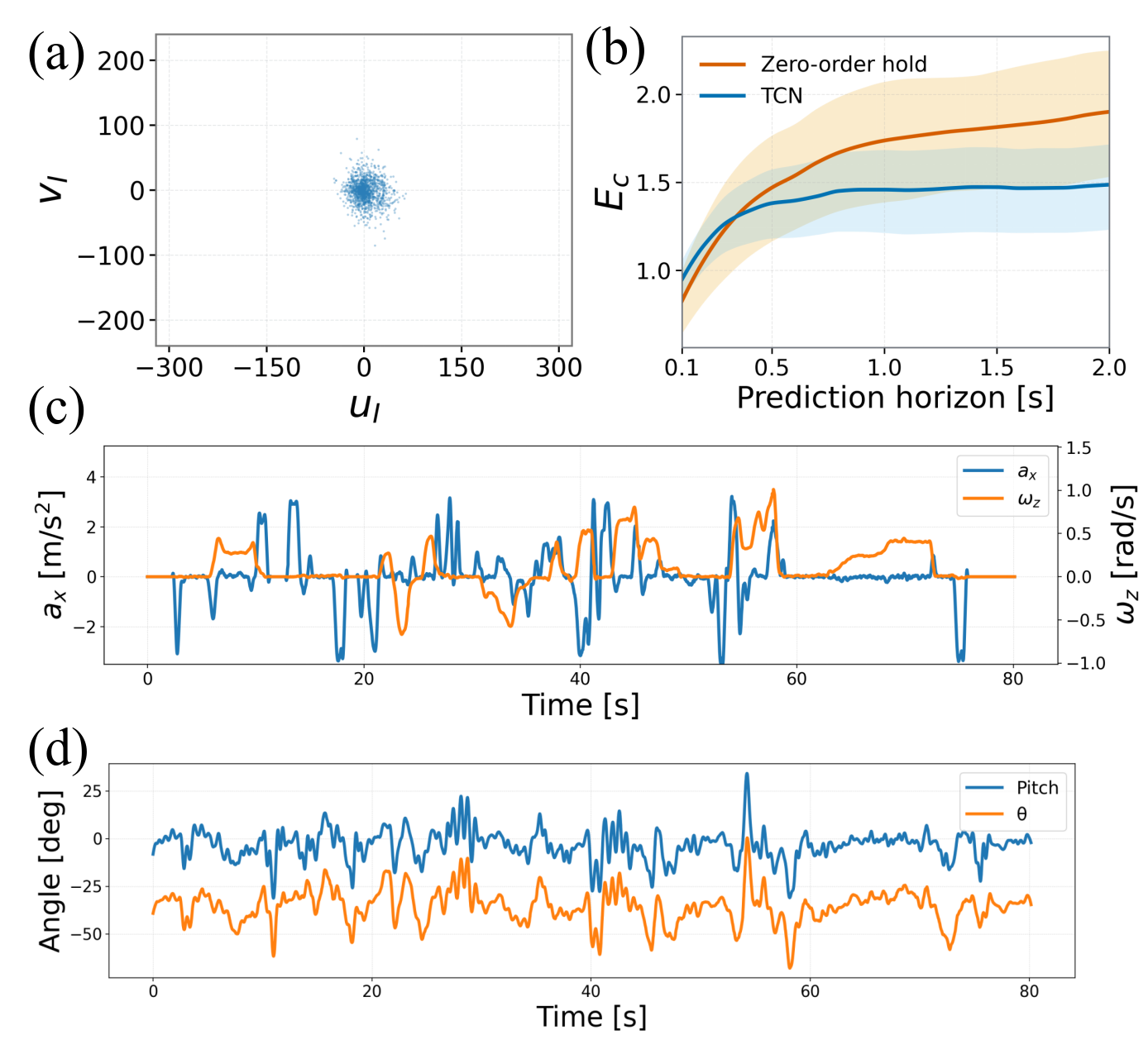}
    \caption{Real-world random-motion experiment. (a) Target image-plane coordinates $(u_I,v_I)$. (b) Combined normalized prediction error $E_c(h)$ for TCN and ZOH; shaded bands denote 95\% block-bootstrap confidence intervals. (c) UGV longitudinal acceleration $a_x$ and yaw rate $\omega_z$. (d) UAV body pitch and gimbal angle $\theta$.}
    \label{fig:realworld_exp2_data}
\end{figure}

\begin{figure}[t]
    \centering
    \includegraphics[width=.88\columnwidth]{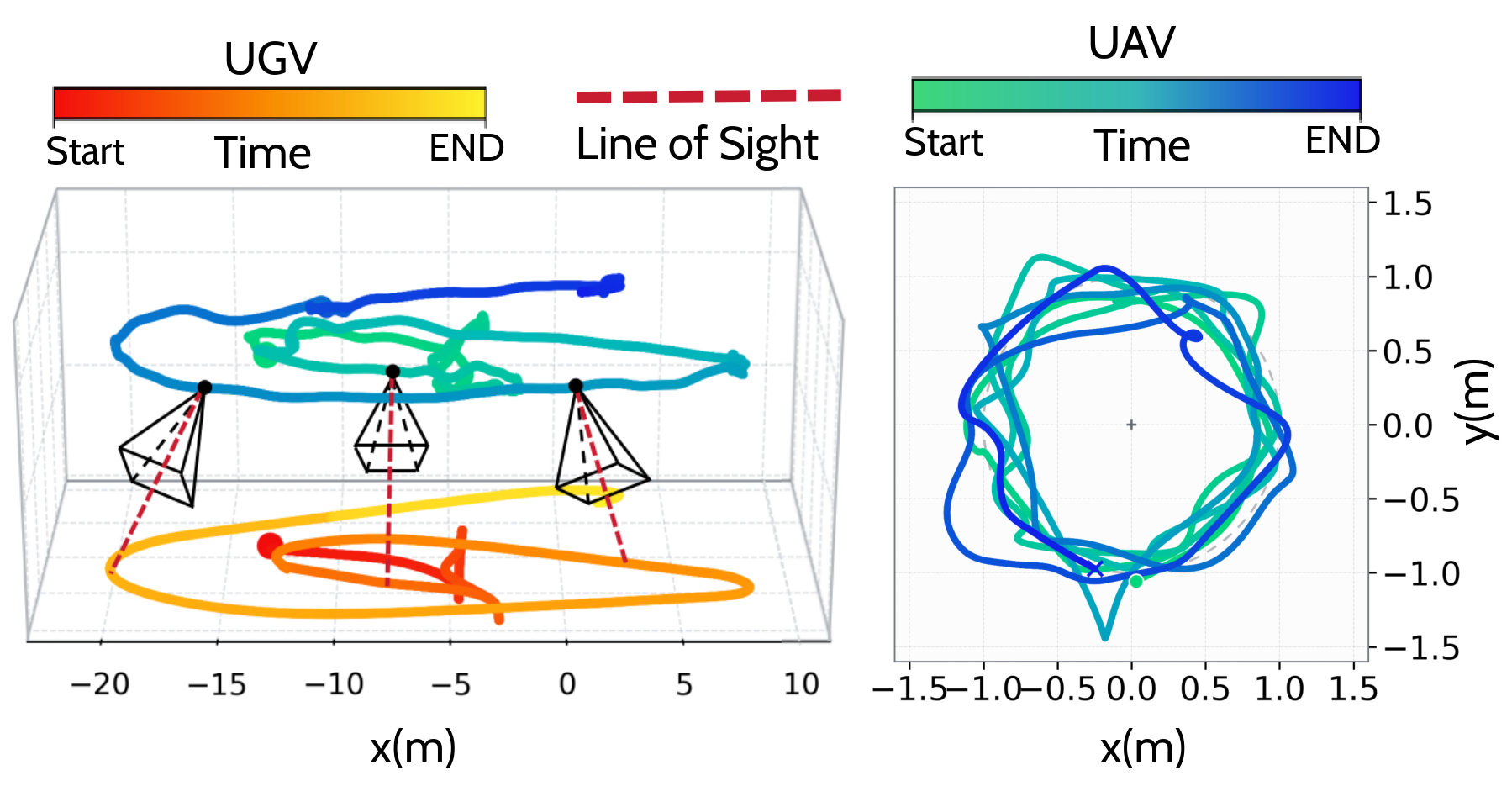}
    \caption{Trajectories in the real-world random-motion experiment. Left: world-frame UGV and UAV trajectories with representative camera poses. Right: UAV trajectory expressed in the UGV frame; the gray dashed circle denotes the desired 1\,m relative orbit.}
    \label{fig:realworld_exp2_trajectory}
\end{figure}

We first evaluate robustness under complex, randomly generated target motion while commanding the UAV to maintain a 1\,m relative circle. The UGV executes abrupt acceleration, deceleration, sharp turns, and repeated S-shaped maneuvers, producing frequent coupled changes in longitudinal acceleration and yaw rate (Fig.~\ref{fig:realworld_exp2_data}(c)). During this trial, the UGV reaches a maximum longitudinal acceleration of $3\,\mathrm{m/s^2}$, a maximum speed of 3\,m/s, and a maximum yaw rate of $1.0\,\mathrm{rad/s}$.

To jointly evaluate the prediction accuracy of the longitudinal acceleration and yaw rate, we define the dimensionless combined normalized error as
\begin{equation}
\begin{aligned}
E_c(h)=\Bigg\{\frac{1}{N_s}\sum_{i=1}^{N_s}\Bigg[
&\left(\frac{\hat a_{x,i}(h)-a_{x,i}(h)}{\sigma_{a_x}}\right)^2 \\
&+\left(\frac{\hat\omega_{z,i}(h)-\omega_{z,i}(h)}{\sigma_{\omega_z}}\right)^2
\Bigg]\Bigg\}^{1/2},
\end{aligned}
\label{eq:combined_prediction_error}
\end{equation}
where $h$ is the prediction lead time, $N_s$ is the number of evaluated prediction samples, and $\sigma_{a_x}$ and $\sigma_{\omega_z}$ are the standard deviations of the corresponding ground-truth signals in the evaluation data. A smaller $E_c$ indicates higher overall prediction accuracy. Fig.~\ref{fig:realworld_exp2_data}(b) shows that the TCN improves long-horizon prediction, lowering $E_c$ by 21.6\% at 2\,s while exhibiting a narrower 95\% confidence interval and hence lower variability.

The gimbal angle $\theta$ compensates for the UAV body-pitch variation in real time (Fig.~\ref{fig:realworld_exp2_data}(d)), keeping the target image-plane positions clustered near the image center (Fig.~\ref{fig:realworld_exp2_data}(a)). The corresponding world-frame and relative trajectories are shown in Fig.~\ref{fig:realworld_exp2_trajectory}. The UAV achieves a measured tracking RMSE of 0.149\,m. These results demonstrate robust tracking under rapidly varying target motion.

\begin{figure*}[!t]
    \centering
    \includegraphics[width=.96\textwidth]{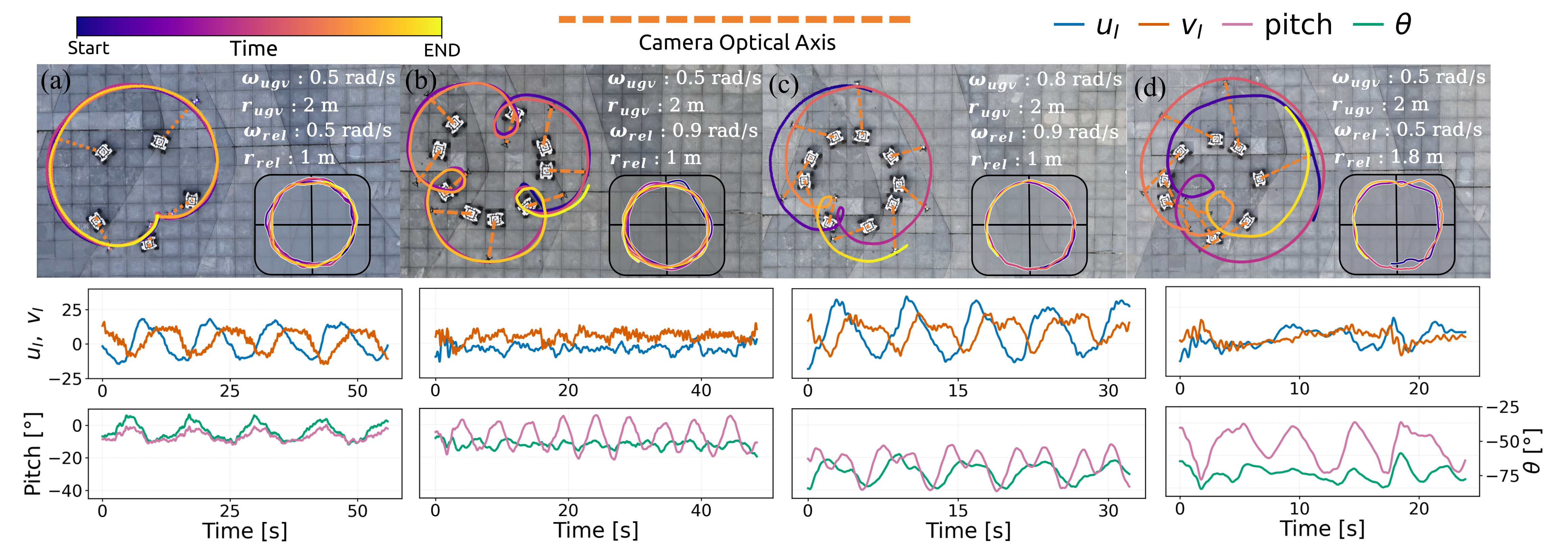}
    \caption{Real-world experiments under four conditions. Each column corresponds to one condition and shows the aerial view with overlaid trajectories, the relative trajectory in the UGV frame, the image-plane errors $u_I$ and $v_I$, and the UAV body pitch and gimbal angle $\theta$ over time, shown on the left and right vertical axes, respectively.}
    \label{fig:realworld_exp1}
\end{figure*}

We then evaluate camera compensation under the four motion conditions shown in Fig.~\ref{fig:realworld_exp1}. Here, $\omega_{\mathrm{ugv}}$ and $r_{\mathrm{ugv}}$ denote the UGV yaw rate and path radius, while $\omega_{\mathrm{rel}}$ and $r_{\mathrm{rel}}$ denote the UAV relative-orbit rate and radius. Case~(b) raises $\omega_{\mathrm{rel}}$ from 0.5 to 0.9\,rad/s; (c) also raises $\omega_{\mathrm{ugv}}$ from 0.5 to 0.8\,rad/s, producing $43^\circ$ body pitch and an $83^\circ$ gimbal angle; and (d) raises $r_{\mathrm{rel}}$ from 1 to 1.8\,m, producing $33^\circ$ pitch and again an $83^\circ$ gimbal angle. Despite the increased compensation, no vertical-FOV exit is observed in the four tested conditions.

These results show that the required camera compensation increases with task difficulty, including higher relative angular velocity, more aggressive UGV rotation, and a larger relative orbit. As the motion becomes more demanding, the gimbal no longer simply follows the UAV body pitch but actively uses a wider angular range, reaching its physical limit in the most challenging cases. Nevertheless, it compensates for large UAV attitude variations, and no target exit from the vertical FOV is observed in the four tested conditions. These results demonstrate that COPA effectively decouples the camera viewing direction from UAV pitch motion and extends the system's operating range under the tested aggressive aerial-ground tracking conditions.

\section{Conclusion}

This paper presents COPA, a robust active-perception framework for global-state-free aerial-ground cooperation. COPA uses a single-axis gimbal to decouple the camera optical axis from UAV pitch, an active-perception model in the non-inertial frame, and a TCN to predict UGV motion. The MPC jointly optimizes UAV and gimbal control. Simulations and real-world experiments demonstrate continuous target visibility, reduced peak tracking errors during motion transitions, and accurate relative tracking under dynamic UGV motion.

\bibliographystyle{IEEEtran}
\bibliography{reference}

\vspace{-0.5cm}

\begin{IEEEbiography}[{\includegraphics
[width=1in,height=1.25in,clip,
keepaspectratio]{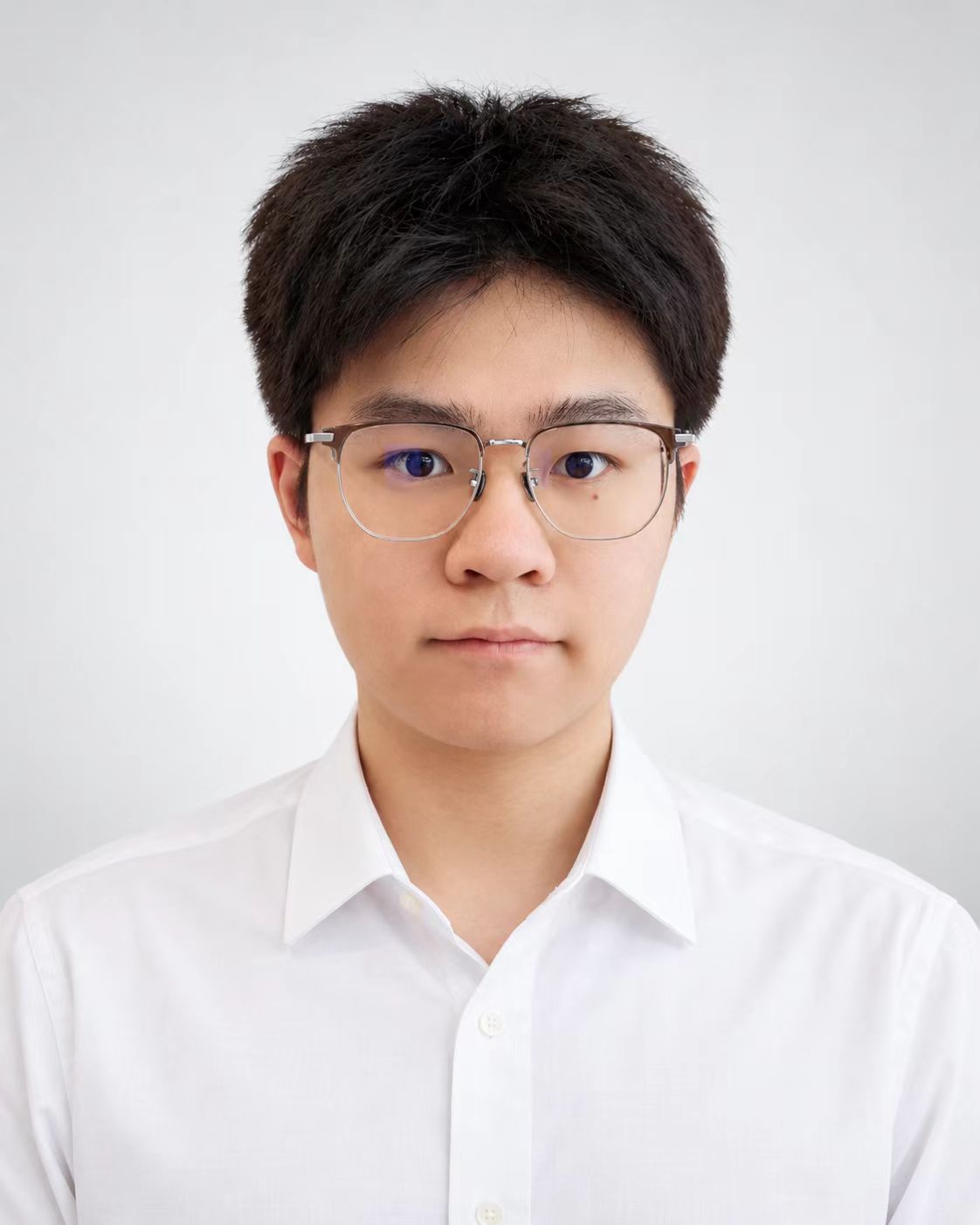}}]
{Mingxuan Zhang} received the B.Eng. degree in automation from Harbin Institute of Technology, Shenzhen, China, in 2024. He is currently pursuing the M.Eng. degree in control science and engineering in the Fast Lab at Zhejiang University, Hangzhou, China. His current research interests include trajectory planning and control, and reinforcement learning.
\end{IEEEbiography}

\vspace{-0.5cm}

\begin{IEEEbiography}[{\includegraphics
[width=1in,height=1.25in,clip,
keepaspectratio]{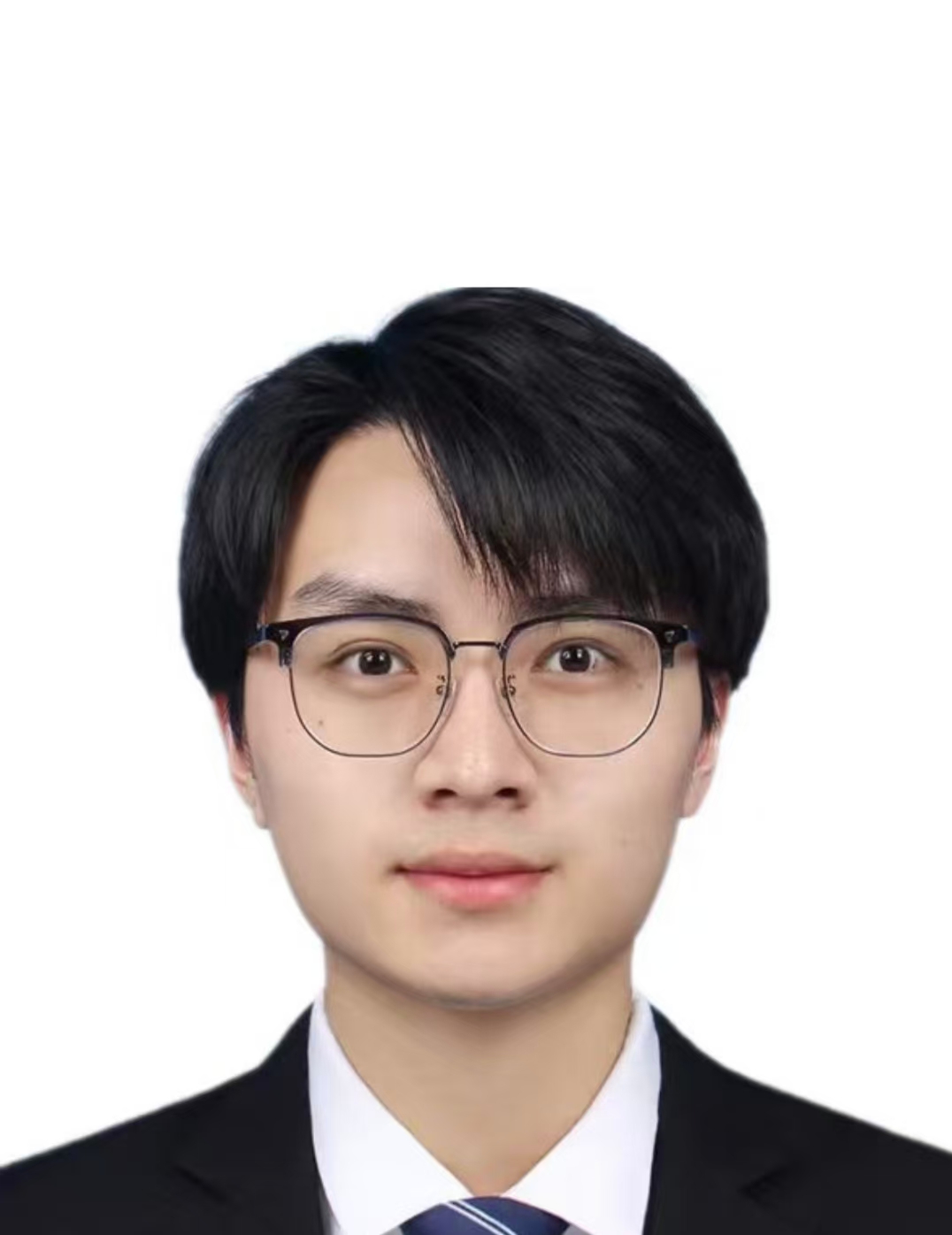}}]
{Jiajun Yu} received the B.E. degree in robotics engineering from Harbin Institute of Technology, Harbin, China, in 2024. He is currently working toward the M.Eng. degree with the College of Control Science and Engineering, Zhejiang University, Hangzhou, China, where he is with the Field Autonomous System and Computing Laboratory (FAST Lab). His research interests include robotic motion planning, parallel optimization, reinforcement learning, and embodied intelligence.
\end{IEEEbiography}

\vspace{-0.5cm}

\begin{IEEEbiography}[{\includegraphics
[width=1in,height=1.25in,clip,
keepaspectratio]{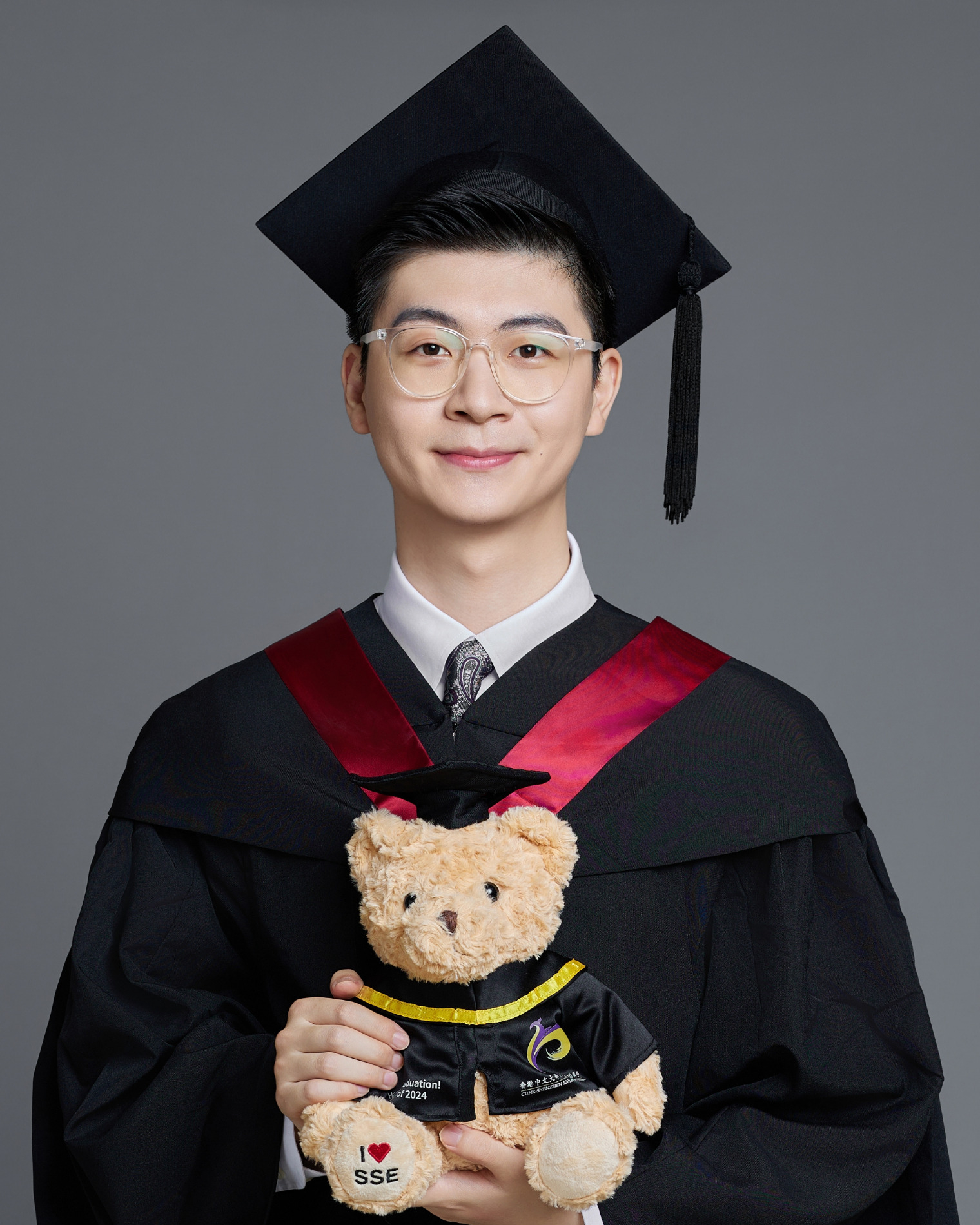}}]
{Baozhe Zhang} received the B.Eng. degree from The Chinese University of Hong Kong, Shenzhen, China, where he is currently pursuing the Ph.D. degree. His research focuses on optimization- and learning-based methods for robot motion planning and control.
\end{IEEEbiography}

\vspace{-0.5cm}

\begin{IEEEbiography}[{\includegraphics
[width=1in,height=1.25in,clip,
keepaspectratio]{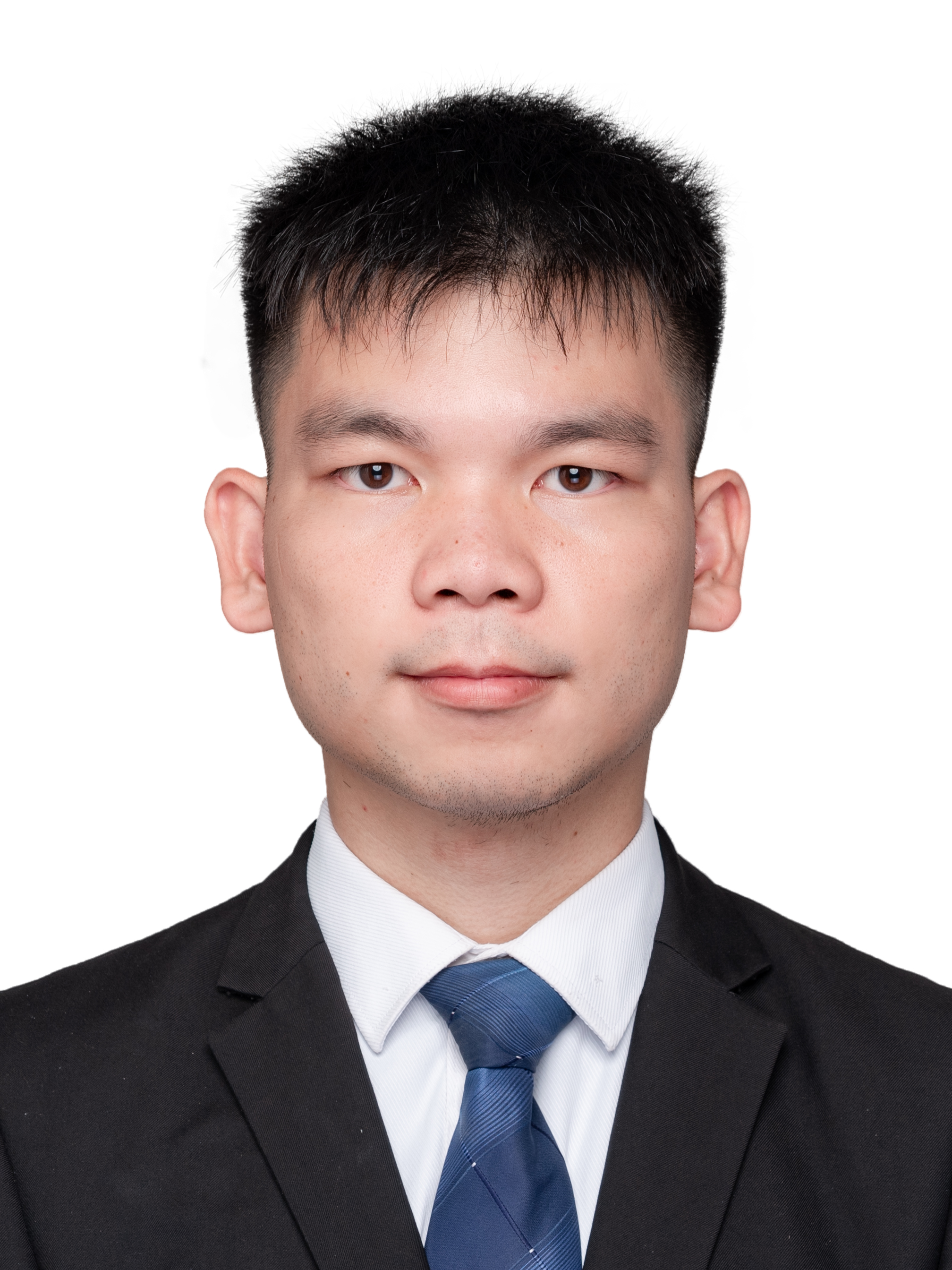}}]
{Pengxiang Zhou} received the B.Eng. degree in electronic information engineering from Southwest Jiaotong University, Chengdu, China, in 2025. He is currently pursuing the M.Eng. degree in electronic information at Zhejiang University, Hangzhou, China. His current research interests include trajectory planning and reinforcement learning.
\end{IEEEbiography}

\vspace{-0.5cm}

\begin{IEEEbiography}[{\includegraphics
[width=1in,height=1.25in,clip,
keepaspectratio]{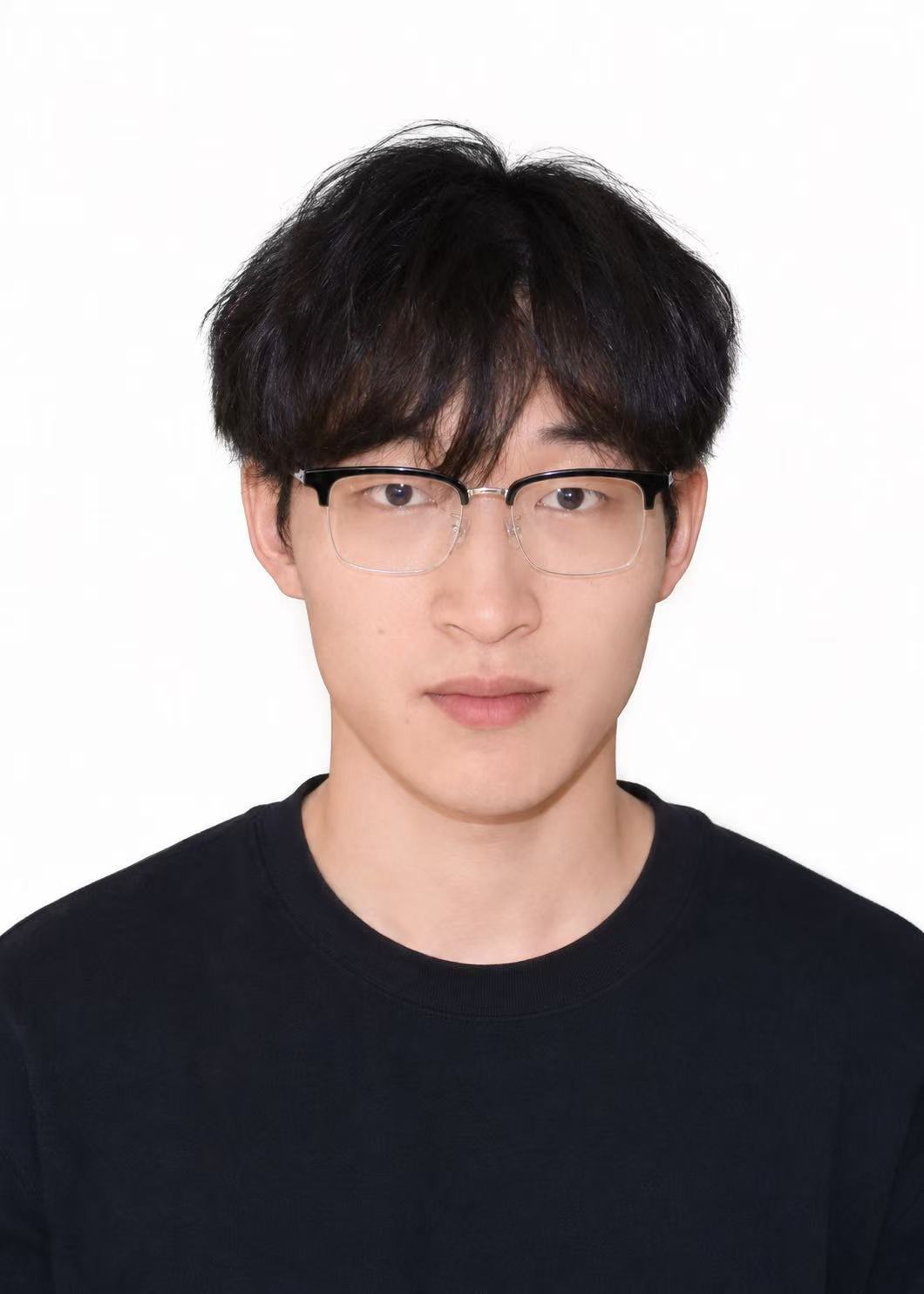}}]
{Wentao Liu} received the B.S. degree in electronic information engineering from Zhejiang University of Technology, Hangzhou, China, in 2024. He is currently pursuing the Ph.D. degree in electronic information at Zhejiang University, Hangzhou, China. His research interests include path planning and trajectory optimization.
\end{IEEEbiography}

\vspace{-0.5cm}

\begin{IEEEbiography}[{\includegraphics
[width=1in,height=1.25in,clip,
keepaspectratio]{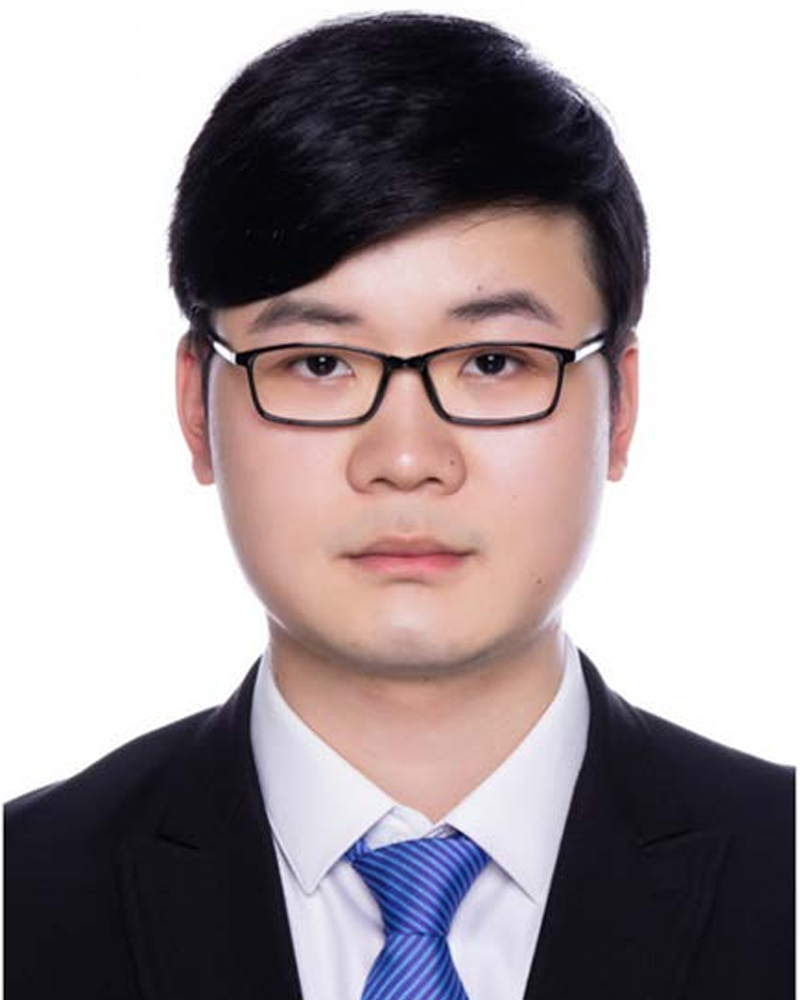}}]
{Fei Gao} received the Ph.D. degree in electronic and computer engineering from the Hong Kong University of Science and Technology, Hong Kong, in 2019. He is currently a tenured associate professor at the Department of Control Science and Engineering, Zhejiang University. His research interests include aerial robots, autonomous navigation, motion planning, optimization, and localization and mapping.
\end{IEEEbiography}

\vspace{-0.5cm}

\begin{IEEEbiography}[{\includegraphics
[width=1in,height=1.25in,clip,
keepaspectratio]{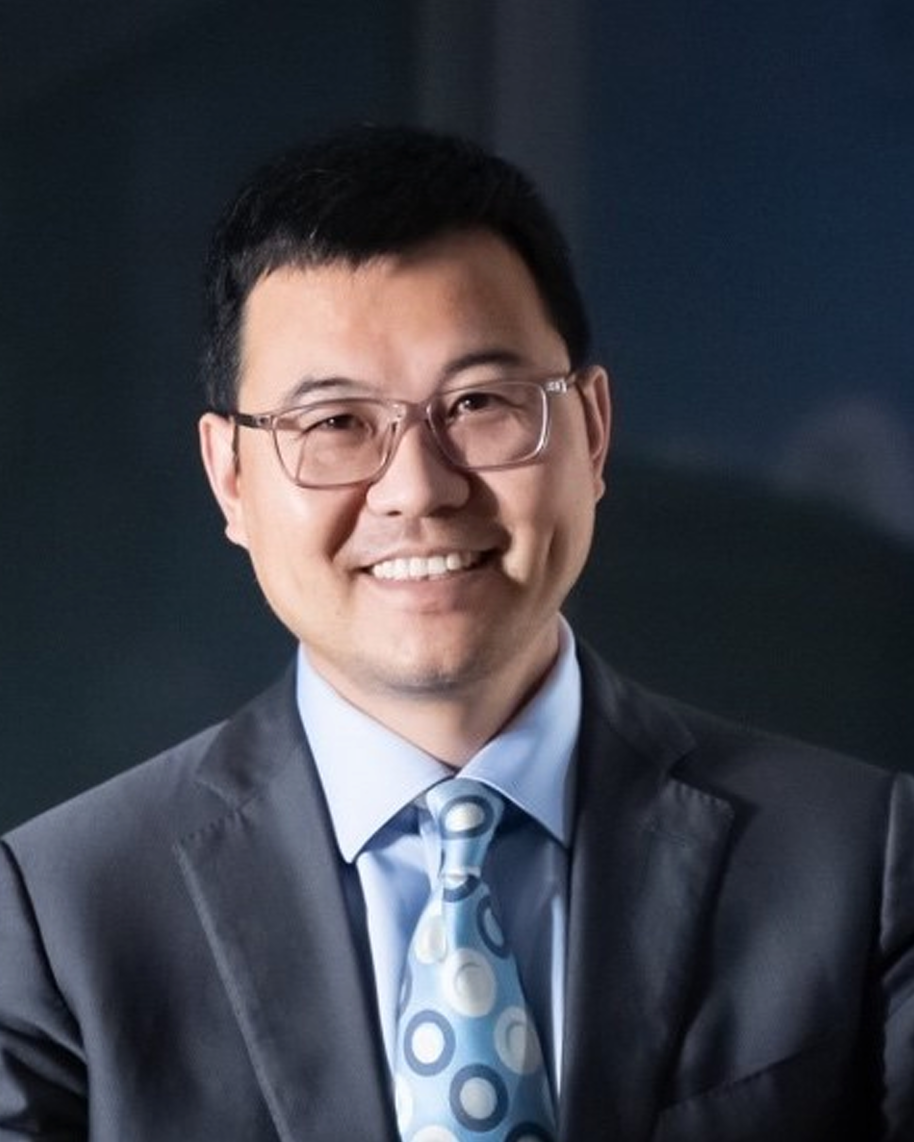}}]
{Chao Xu} received the Ph.D. degree in mechanical engineering from Lehigh University in 2010. He is currently the Associate Dean and a Professor with the College of Control Science and Engineering, Zhejiang University. He is the Inaugural Dean of ZJU Huzhou Institute. His research expertise is flying robotics and control-theoretic learning. He has published over 100 articles in international journals, including Science Robotics and Nature Machine Intelligence. He will join the organization committee of the IROS-2025 in Hangzhou.
\end{IEEEbiography}

\vspace{-0.5cm}

\begin{IEEEbiography}[{\includegraphics
[width=1in,height=1.25in,clip,
keepaspectratio]{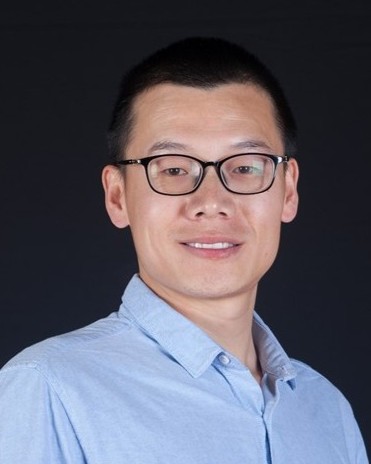}}]
{Yanjun Cao} received his Ph.D. degree in computer and software engineering from the University of Montreal, Polytechnique Montreal, Canada, in 2020. He is currently an associate researcher at the Huzhou Institute of Zhejiang University, as a PI in the Center of Swarm Navigation. He leads the Field Intelligent Robotics Engineering group of the Field Autonomous System and Computing Lab. His research focuses on key challenges in multi-robot systems, such as collaborative localization, autonomous navigation, perception and communication.
\end{IEEEbiography}

\end{document}